\documentclass{svjour3} 

\smartqed 
\usepackage{graphicx}
\usepackage{url}
\usepackage[hidelinks]{hyperref}
\usepackage{booktabs, tabularx, threeparttable, xfrac, algorithm, array, multirow, amsmath, amssymb}
\usepackage{natbib}
\usepackage[justification=centering]{caption}

\usepackage{tikz}
\def\checkmark{\tikz\fill[scale=0.4](0,.35) -- (.25,0) -- (1,.7) -- (.25,.15) -- cycle;}

\definecolor{ceruleanblue}{rgb}{0.16, 0.32, 0.75}

\newcolumntype{Y}{>{\centering\arraybackslash}X}
\newcolumntype{C}[1]{>{\centering\arraybackslash}m{#1}}

\usepackage[noend]{algpseudocode}

\algrenewcommand\textproc{}
\makeatletter
\newcommand{\multiline}[1]{%
	\begin{tabularx}{\dimexpr\linewidth-\ALG@thistlm}[t]{@{}X@{}}
		#1
	\end{tabularx}
}
\makeatother

\journalname{This is the author’s version of an article published in Data Mining and}

\begin{document}

\title{XEM: An Explainable-by-Design Ensemble Method\\ for Multivariate Time Series Classification}
\titlerunning{XEM: An Explainable-by-Design Ensemble Method for MTS Classification}

\author{Kevin Fauvel$^1$         \and
        \'Elisa Fromont$^2$		 \and
        V\'eronique Masson$^1$	 \and
        Philippe Faverdin$^3$	 \and 
        Alexandre Termier$^1$       
}

\authorrunning{Kevin Fauvel et al.}

\institute{$^1$Inria, Univ Rennes, CNRS, IRISA, France\\
		   $^2$Univ Rennes, IUF, Inria, CNRS, IRISA, France\\
           $^3$PEGASE, INRAE, AGROCAMPUS OUEST, France
}

\date{}

\maketitle

\begin{abstract}
We present XEM, an eXplainable-by-design Ensemble method for Multivariate time series classification. XEM relies on a new hybrid ensemble method that combines an explicit boosting-bagging approach to handle the bias-variance trade-off faced by machine learning models and an implicit divide-and-conquer approach to individualize classifier errors on different parts of the training data. Our evaluation shows that XEM outperforms the state-of-the-art MTS classifiers on the public UEA datasets. Furthermore, XEM provides faithful explainability-by-design and manifests robust performance when faced with challenges arising from continuous data collection (different MTS length, missing data and noise).

\end{abstract}

\section{Introduction}
The prevalent deployment and usage of sensors in a wide range of sectors generate an abundance of multivariate data which has been proven to be instrumental for researches, businesses and policies~\citep{Esteva19,Ransbotham19,Cussins19}. In particular, multivariate data that integrates temporal evolution has received significant interests over the past decade, driven by automatic and high-resolution monitoring applications (e.g., healthcare~\citep{Li18}, mobility~\citep{Jiang19}, natural disasters~\citep{Fauvel20}).

In our study, we focus on the issue of multivariate data classification, which consists of learning the relationship between a multivariate sample and its label. Specifically, we study the Multivariate Time Series (MTS) classification setting. A time series is a sequence of real values ordered according to time; and when a set of coevolving time series are recorded simultaneously by a set of sensors, it is called an MTS.

In addition to prediction performance, machine learning methods have to be assessed on how they can support their predictions with explanations in many cases (e.g., decision support, legal requirement, model validation). 
In particular, machine learning methods have to be assessed on how they can provide faithful explanations.
Faithfulness is critical as it corresponds to the level of trust an end-user can have in the explanations of model predictions, i.e. the level of relatedness of the explanations to what the model actually computes.
The best performing state-of-the-art MTS classifiers on the public UEA archive~\citep{Bagnall18} are ``black-box'' models (MLSTM-FCN~\citep{Karim19}, WEASEL+MUSE~\citep{Schafer17}), i.e. complicated-to-understand models~\citep{Lipton16}. Nonetheless, black-box models like MLSTM-FCN and WEASEL+MUSE cannot support their predictions with faithful explanations as they can only rely on explainability methods providing explanations from any machine learning model~\citep{Rudin19} (post hoc model-agnostic explainability methods).
Therefore, we propose a new MTS classifier that combines performance and faithful explainability.
Our new approach generates features which enable it to outperform the state-of-the-art MTS classifiers on the UEA datasets, while providing faithful explainability-by-design through identifying the time window used to classify the whole MTS.

Some feature-based MTS classifiers exist in the state-of-the-art (gRSF \citep{Karlsson16}, LPS~\citep{Baydogan16}, mv-ARF~\citep{Tuncel18}, SMTS~\citep{Baydogan14} and WEASEL+ MUSE \citep{Schafer17}).
However, the features generated by these MTS classifiers cannot be used as explanations to support the models' predictions as they do not allow, by design, the identification of the regions of the input data that are important for predictions. First, the shapelet-based approach gRSF creates a black-box classifier (a forest of decision trees) over randomly extracted subsequences (shapelets), which prevents the direct extraction from the model of shapelets important for predictions.
Then, the bag-of-words approaches (LPS, mv-ARF, SMTS, WEASEL+MUSE) convert time series into a bag of discrete words, and use a histogram of words representation to perform the classification. The bag of discrete words generated by these approaches (symbolic representations from decision trees predictions, unigrams/bigrams extraction following a Symbolic Fourier Approximation~\citep{Schafer12}) are difficult to understand and cannot be mapped to the regions of the input data that are important for predictions.
Therefore, we propose a new MTS classifier that generates features allowing the direct identification of the MTS time window that is important for prediction. These features correspond to the confidence levels of a classifier on each MTS subsequence of a predefined length. The subsequence where the classifier is the most confident is used for classification and provided to the end-user as faithful explanation to support the MTS prediction.
Thus, our new MTS classifier relies on the development of a well-performing classifier that is applied to MTS subsequences. 
As in~\citep{Baydogan14}, we have chosen a tabular classifier because it fulfills two needs simultaneously: first, the need to handle the relationship between the variables; second, the need to handle really small time series according to the predefined time window length of interest (e.g., time series length of 2). Most MTS classifiers fail to meet the second need. 

To undertake the task of the tabular multivariate classification, no single classifier can claim to be superior to any of the others~\citep{Wolpert96} (known as the ``No Free Lunch theorem"). Thus, the combination of different classifiers - an ensemble method - is often considered a good method to obtain a better generalizing classifier. There are three main reasons that justify the use of ensembles over single classifiers~\citep{Dietterich00}: statistical (reduce the risk of choosing the wrong classifier by averaging when the amount of training data available is too small compared to the size of the hypothesis space), computational (local search from many different starting points may provide a better approximation to the true unknown function than any of the individual classifier), and representational (expansion of the space of representable functions). 

The construction of an ensemble method involves combining accurate and diverse individual classifiers. 
There are two complementary ways to generate diverse classifiers. 
First, each individual classifier can be set to learn a different part of the original training data~\citep{Masoudnia14}. 
For example, Local Cascade (LC)~\citep{Gama00} is a state-of-the-art method adopting this first diversification way.
LC learns different part of the training data to capture new relationships that cannot be discovered globally based on a divide-and-conquer strategy (a decision tree). Then, LC manages the bias-variance trade-off faced by machine learning models through the use, at each level of the tree, of classifiers with different behaviors.
However, methods relying on learning different parts of the training data like LC do not benefit from the second diversification way, which consists of generating classifiers by perturbing the distribution of the original training data~\citep{Sharkey97}. Sharkey et al. argued that training classifiers using different training sets produces low correlated errors. Within this way, there are two well-known methods that modify the distribution of the original training data with complementary effects on the bias-variance trade-off: bagging~\citep{Breiman96} (variance reduction) and boosting~\citep{Schapire99} (bias reduction). We call an ensemble method which fully adopts these two ways to generate diverse classifiers a hybrid ensemble method. As far as we have seen, we have developed in~\citep{Fauvel19} the \emph{first} hybrid ensemble method (Local Cascade Ensemble - LCE).
The new hybrid ensemble method combines a boosting-bagging approach to handle the bias-variance trade-off (second diversification way) and, as LC, a divide-and-conquer approach - a decision tree - to learn different parts of the training data (first diversification way).

However,~\citep{Fauvel19} does not show how LCE behaves on public tabular multivariate datasets (e.g., UCI repository) since it was only applied to a proprietary dataset. 
Therefore, in this paper, we first present in detail and thoroughly examine the behavior of LCE.
Then, we present how LCE is used to form an eXplainable Ensemble method for MTS classification (XEM) combining both performance and faithful explainability.
Finally, we highlight some interesting properties of XEM, and in particular that XEM is robust with varying MTS input data quality (different MTS length, missing data and noise), which often arises in continuous data collection systems.
Summarizing our main contributions:
\begin{itemize}
	\item[$\bullet$] We detail the presentation of LCE algorithm introduced in~\citep{Fauvel19}, in particular with regard to its properties and time complexity;
	\item[$\bullet$] We examine the behavior of LCE on a public benchmark, as LCE was only applied to a proprietary dataset in~\citep{Fauvel19}. Our study shows that LCE outperforms the state-of-the-art tabular classifiers on the public UCI datasets~\citep{Dua17};
	\item[$\bullet$] Leveraging LCE, we present a new eXplainable Ensemble method for MTS classification (XEM) combining performance and faithful explainability. XEM outperforms the state-of-the-art MTS classifiers on the public UEA datasets~\citep{Bagnall18} and provides faithful explainability-by-design through identifying the time window used to classify the whole MTS;
	\item[$\bullet$] We show that XEM manifests robust performance when faced with challenges arising from continuous data collection (different MTS length, missing data and noise).
\end{itemize}

The rest of this paper is organized as follows: Section~\ref{RelatedWork} presents the related work concerning classification, MTS classification and explainability; Section~\ref{Algorithm} details LCE and XEM; Section~\ref{Evaluation} presents our evaluation method; and finally, Section~\ref{Results} discusses our results.

\vspace{-1em}
\section{Background and Related Work}
\label{RelatedWork}
In this section we first introduce the background of our study. Then, we present the state-of-the-art tabular classification methods on which we position our algorithm LCE, and we end with a similar presentation for MTS classification.

\vspace{-1em}
\subsection{Background}
We address the issue of supervised learning for classification. \textit{Classification} consists of learning a function that maps an input data to its label: given an input space $X$, an~output space $Y$, an~unknown distribution ${P}$ over $X \times Y$, a~training set sampled from ${P}$, and a~0--1 loss function $\ell_{0-1}$ compute function $h^*$ as follows:

\begin{equation}
	{h^* = \underset{h}{\arg\min} \quad \mathbb{E}_{(x,y) \sim P} \left[ \ell_{0-1}(h, (x,y)) \right]}
\end{equation}
\vspace{-.5em}

Our classifier LCE is based on a new way to handle the bias-variance trade-off in ensemble methods. The bias-variance trade-off defines the capacity of the learning algorithm to generalize beyond the training set. The \textit{bias} is the component of the classification error that results from systematic errors of the learning algorithm. A high bias means that the learning algorithm is not able to capture the underlying structure of the training set (underfitting). The \textit{variance} measures the sensitivity of the learning algorithm to changes in the training set. A high variance means that the algorithm is learning too closely the training set (overfitting). The objective is to minimize both the bias and variance.

We perform classification on two types of datasets: traditional (tabular) multivariate data and MTS. In the traditional multivariate data setting, in contrast to the MTS one, there is no explicit relationship among samples or variables and every sample has the same set of variables (also called attributes or dimensions). A \textit{Multivariate Time Series} (MTS) $M=\{x_1,...,x_d\} \in \mathcal{R}^{d*l}$ is an ordered sequence of $d \in \mathcal{N}$ streams with $x_i=(x_{i,1},...,x_{i,l})$, where $l$ is the length of the time series and $d$ is the number of multivariate dimensions. We address MTS generated from automatic sensors with a fixed and synchronized sampling along all dimensions. An example of an MTS dataset is given at the top of Figure~\ref{fig:Dataset_Transformation}. This dataset contains n MTS with 2 dimensions and a time series length of 5.

\vspace{-1em}
\subsection{Classification}
\label{RW_Classification}
In machine learning, the most popular (and often best performing) classifiers belong to the following classes: regularized logistic regressions, support vector machines, neural networks and ensemble methods. As previously discussed, ensemble methods are usually well generalizing classifiers and thus, we position our approach into this class. The other classes constitute our competitors and the algorithms evaluated are presented in section~\ref{Classifiers_tabular}.

Ensemble methods are structured around two approaches (explicit, implicit) which have their own strengths and limitations. Therefore a hybrid ensemble method is encouraged~\citep{Masoudnia14}. 
The \textit{implicit approach} involves creating diverse classifiers on the original training data, whereas the \textit{explicit approach} emphasizes classifiers diversity through the creation of different training sets by probabilistically changing the distribution of the original training data.

There are two methods adopting an implicit approach: Mixture of Experts (ME)~\citep{Jacobs91} and Negative Correlation Learning (NCL)~\citep{Liu99}. \textit{ME} uses a divide-and-conquer algorithm to split the problem space, and each individual classifier learns a part of the training data. The advantage of this method is that each individual classifier is concerned with its own individual error. However, individual classifiers are trained independently so there is no control over the bias-variance trade-off. 
Next, \textit{NCL} is an ensemble method which is trained on the entire training data simultaneously and interactively to adjust the bias-variance trade-off. Individual classifiers interact through the correlation penalty terms of their error functions. The correlation penalty term is a regularization term that is integrated into the error function of each individual classifier. This term quantifies the amount of error correlation and is minimized during the training, which leads to negatively correlated individual classifiers and balances the bias-variance trade-off. The disadvantage of this method is that each classifier is concerned with the whole ensemble error due to the training of each classifier on the same data. Some studies like Local Cascade~\citep{Gama00} combine NCL and ME features to address their limitations.

However, a combination of implicit approaches does not benefit from the diversification of generating classifiers by perturbing the distribution of the original training data (explicit approach). There are two methods adopting an explicit approach with complementary effects on the bias-variance trade-off (bagging~\citep{Breiman96} - variance reduction, boosting~\citep{Schapire99} - bias reduction). \textit{Bagging} is a method for generating multiple versions of a predictor (bootstrap replicates) and using these to get an aggregated predictor. \textit{Boosting} is a method for iteratively learning weak classifiers and adding them to create a final strong classifier. After a weak learner is added, the data weights are readjusted, allowing future weak learners to focus more on the examples that previous weak learners misclassified. Bagging and boosting methods have been combined~\citep{Kotsiantis05} but without integrating the diversification benefit of an implicit approach.

There is a study which combines the explicit boosting method with the implicit ME divide-and-conquer principle~\citep{Ebrahimpour12}. Nonetheless, the only bias reduction distribution change of boosting does not ensure a bias-variance trade-off. Hence, we propose the first hybrid ensemble method called Local Cascade Ensemble (LCE). LCE combines an explicit boosting-bagging approach to handle the bias-variance trade-off and an implicit divide-and-conquer approach (decision tree) to learn different parts of the training data. 

Therefore, in this work we choose to evaluate the performance of the first hybrid ensemble method LCE in comparison to:
\begin{itemize}	
	\item[$\bullet$] A simple ensemble method on the original data combining some state-of-the-art classifiers with a majority vote (Na\"ive Bayes~\citep{Zhang04}, Elastic Net~\citep{Zou05}, CART~\citep{Breiman84});
	\item[$\bullet$] The state-of-the-art ensemble methods adopting an explicit approach (Random Forest~\citep{Breiman01}, Extreme Gradient Boosting~\citep{Chen16} and the combination of bagging and boosting~\citep{Kotsiantis05});
	\item[$\bullet$] The state-of-the-art implicit approach Local Cascade~\citep{Gama00} (starting point of LCE - detailed in section~\ref{Algorithm});
	\item[$\bullet$] The state-of-the-art ensemble method combining the explicit boosting method with the implicit ME divide-and-conquer principle (Boost-Wise Pre-Loaded Mixture of Experts~\citep{Ebrahimpour12});
	\item[$\bullet$] The best-in-class of the other classes (regularized logistic regressions, support vector machines and neural networks) as presented in section~\ref{Classifiers_tabular}.
\end{itemize}

\vspace{-1em}
\subsection{MTS Classification}
\label{RW_MTSClassification}

\textbf{MTS Classifiers}
We can categorize the state-of-the-art MTS classifiers into three families: similarity-based, feature-based and deep learning methods.

Similarity-based methods make use of similarity measures (e.g., Euclidean distance) to compare two MTS. Dynamic Time Warping (DTW) has been shown to be the best similarity measure to use along k-NN~\citep{Seto15}, this approach is called kNN-DTW. There are two versions of kNN-DTW for MTS: dependent (DTW$_{D}$) and independent (DTW$_{I}$). Neither dominates over the other~\citep{Shokoohi17}. DTW$_{I}$ measures the cumulative distances of all dimensions independently measured under DTW. DTW$_{D}$ uses a similar calculation with single-dimensional time series; it considers the squared Euclidean cumulated distance over the multiple dimensions.

Feature-based methods include shapelets and bag-of-words (BoW) models. Shapelets models use subsequences (shapelets) to transform the original time series into a lower-dimensional space that is easier to classify. gRSF~\citep{Karlsson16} and UFS~\citep{Wistuba15} are the current state-of-the-art shapelets models in MTS classification. They relax the major limiting factor of the time to find discriminative subsequences in multiple dimensions (shapelet discovery) by randomly selecting shapelets. gRSF creates decision trees over randomly extracted shapelets and shows better performance than UFS on average (14 MTS datasets)~\citep{Karlsson16}. On the other hand, BoW models (LPS~\citep{Baydogan16}, mv-ARF~\citep{Tuncel18}, SMTS~\citep{Baydogan14} and WEASEL+MUSE~\citep{Schafer17}) convert time series into a bag of discrete words, and use a histogram of words representation to perform the classification. WEASEL+MUSE shows better results compared to gRSF, LPS, mv-ARF and SMTS on average (20 MTS datasets)~\citep{Schafer17}. WEASEL+MUSE generates a BoW representation by applying various sliding windows with different sizes on each discretized dimension (Symbolic Fourier Approximation~\citep{Schafer12}) to capture features (unigrams, bigrams, dimension idenfication). Following a feature selection with chi-square test, it classifies the MTS based on a logistic regression.

Finally, deep learning methods (FCN~\citep{Wang17}, MLSTM-FCN \citep{Karim19}, ResNet~\citep{He16}, TapNet~\citep{Zhang20} and TST~\citep{Zerveas21}) use Long-Short Term Memory (LSTM), Convolutional Neural Networks (CNN) or Transformers. According to the results published and our experiments, the current state-of-the-art model (MLSTM-FCN) is proposed in~\citep{Karim19} and consists of a LSTM layer and a stacked CNN layer along with squeeze-and-excitation blocks to generate latent features.
A recent network, TapNet~\citep{Zhang20}, also consists of a LSTM layer and a stacked CNN layer, followed by an attentional prototype network. However, TapNet shows lower accuracy results\footnote{\url{https://github.com/xuczhang/xuczhang.github.io/blob/master/papers/aaai20_tapnet_full.pdf}} on average on the 30 public UEA MTS datasets than MLSTM-FCN (MLSTM-FCN results presented in Table~\ref{tab:UEA}).

Therefore, in this work we choose to evaluate the performance of XEM in comparison to the similarity-based methods results published in the UEA archive (ED, DTW$_{D}$, DTW$_{I}$)~\citep{Bagnall18} and to the best-in-class for each feature-based and deep learning category (WEASEL+MUSE and MLSTM-FCN classifiers). As a method aggregating features which are the output of multiple predictors, XEM can be categorized as an ensemble method.

As previously introduced, in addition to meeting the performance requirement, MTS classifiers are facing two particular challenges: the lack of faithful explainability supporting their predictions and the varying input data quality (different TS length, missing data, noise). 

\textbf{Explainability} 
There is no mathematical definition of explainability. A definition proposed by~\citep{Miller19} states that the higher the explainability of a machine learning algorithm, the easier it is for someone to comprehend why certain decisions or predictions have been made.
Three categories of explainability methods are usually recognized: explainability-by-design, post hoc model-specific explainability and post hoc model-agnostic explainability~\citep{Du20}.
First, some machine learning models provide explainability-by-design. These self-explanatory models incorporate explainability directly to their structures. This category includes, for example, decision trees, rule-based models and linear models.
Next, post hoc model-specific explainability methods are specifically designed to extract explanations for a particular model. These methods usually derive explanations by examining internal model structures and parameters. For example, a method has been designed to identify the regions of input data that are important for predictions in CNNs using the class-specific gradient information~\citep{Selvaraju19}.
Finally, post hoc model-agnostic explainability methods provide explanations from any machine learning model. These methods treat the model as a black-box and does not inspect internal model parameters. The main line of work consists in approximating the decision surface of a model using an explainable one (e.g., LIME~\citep{Ribeiro16}, SHAP~\citep{Lundberg17}, Anchors~\citep{Ribeiro18}, LORE~\citep{Guidotti19}). 
These different explainability methods come with their own form of explanations.
Therefore, we have proposed in~\citep{Fauvel20Framework} a framework to assess and benchmark machine learning methods with respect to their performance and explainability. The framework details a set of characteristics (performance, model comprehensibility, granularity of the explanations, information type, faithfulness and user category) that systematize the performance-explainability assessment of machine learning methods.
According to this framework, none of the state-of-the-art MTS classifiers reconciles performance and faithful explainability.
Similarity-based methods provide faithful explainability-by-design but they are often less accurate than other MTS classification methods.
WEASEL+MUSE and MLSTM-FCN classifiers show better performance than similarity-based methods but they are not explainable-by-design and, as far as we have seen, they do not have a post hoc model-specific explainability method. 
Thus, WEASEL+MUSE and MLSTM-FCN cannot provide faithful explanations as they can only rely on post hoc model-agnostic explainability methods~\citep{Rudin19}, which could prevent their use on numerous applications.
Our approach XEM proposes to reconcile performance and faithful explainability (by design) through identifying the time window used to classify the whole MTS. We detail the assessment of XEM in the performance-explainability framework and identify ways to further enhance XEM explainability in section~\ref{framework}.

\textbf{Input Data Quality}
Finally, none of the state-of-the-art MTS classifiers handles the three varying data quality aspects (different TS length, missing data, noise). 

Table~\ref{tab:MTSClassifiers} presents an overview of the challenges addressed by the state-of-the-art MTS classifiers and how we position our new ensemble method XEM. We evaluate the classification performance of XEM and its ability to handle the challenges MTS classification faces in section~\ref{XEM_Evaluation}.

\begin{table}[!htpb]
	\caption{Overview of the state-of-the-art MTS classifiers.}
	\label{tab:MTSClassifiers}
	\centering
	\scriptsize
	\begin{tabularx}{.96\linewidth}{m{3.3cm}>{\centering}m{.6cm}>{\centering}m{.6cm}>{\centering}m{1.5cm}>{\centering}m{1.5cm}|>{\centering\arraybackslash}m{1.5cm}}
		\toprule
		\textbf{}&\multicolumn{2}{C{1.2cm}}{\textbf{Similarity Based}}&\textbf{Deep Learning}&\textbf{Feature Based}&\textbf{Ensemble}\\
		\cline{2-6}\\
		&\textbf{ED}&\textbf{DTW}&\textbf{MLSTM FCN}&\textbf{WEASEL+ MUSE}&\textbf{XEM}\\
		\midrule\midrule[.1em]
		\textbf{Output} & & & & &\\
		\hspace{.4cm}Performance & & & \checkmark & \checkmark & \checkmark \\
		\hspace{.4cm}Faithful Explainability & \checkmark & \checkmark & & & \checkmark\\
		& & & & &\\
		\textbf{Input} & & & & &\\
		\hspace{.4cm}Varying TS Length & & \checkmark & \checkmark & \checkmark & \checkmark\\
		\hspace{.4cm}Missing Data & & & & & \checkmark\\
		\hspace{.4cm}Noise & & & & \checkmark & \checkmark\\
		\bottomrule
	\end{tabularx}
\end{table}

\vspace{-2em}
\section{Algorithm}
\label{Algorithm}
We first explain how the hybrid ensemble method LCE has been designed and then how LCE is used to form the MTS classifier XEM. Finally, we detail XEM properties and implementation.

\vspace{-1em}
\subsection{LCE}
\label{LCEPresentation}
\begin{figure*}[!htpb]
	\centering
	\includegraphics[width=\linewidth]{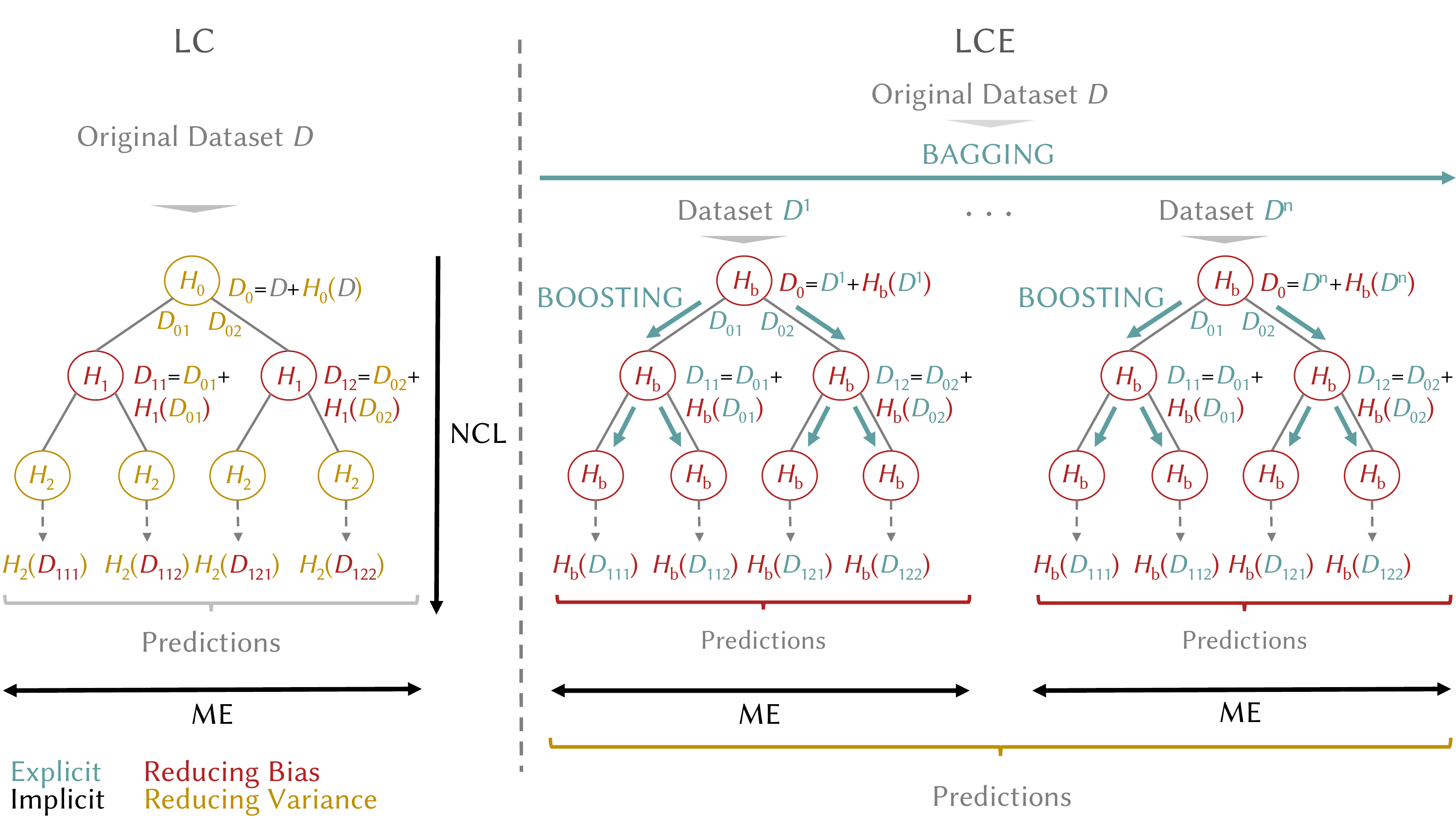}
	\caption{Local Cascade (LC) versus Local Cascade Ensemble (LCE).\\ $H_i$ - base classifier trained on a dataset at a tree depth of i ($H_b$: eXtreme Gradient Boosting [Chen et al., 2016]), $D_i$ - dataset at a tree depth of i augmented with the class probabilities of the base classifier $H_i$, NCL - Negative Correlation Learning, ME - Mixture of Experts.}
	\label{fig:LCE}
\end{figure*}

First of all, LCE is an improved hybrid (explicit and implicit) version of an implicit cascade generalization approach~\citep{Sesmero15}: Local Cascade (LC)~\citep{Gama00}. Among the implicit approaches, LC is one of the easiest to augment with explicit techniques. LC uses a decision tree as a divide-and-conquer method, which is compatible with the explicit bagging/boosting approaches. This criteria has motivated the choice of LC algorithm as the starting point for our hybrid ensemble method. We present in this section LC and our proposed LCE. Figure~\ref{fig:LCE} illustrates the algorithms.

\vspace{-1em}
\vspace{-1em}

\textbf{Local Cascade} LC is a combined implicit approach (negative correlation learning and mixture of experts) based on a cascade generalization. Cascade generalization uses a set of classifiers sequentially and at each stage adds new attributes to the original dataset. The new attributes are derived from the class probabilities given by a classifier, called a base classifier (e.g., class probabilities $H_{0}(D)$, $H_{1}(D_{01})$ in Figure~\ref{fig:LCE}). The bias-variance trade-off is obtained by negative correlation learning: at each stage of the sequence, classifiers with different behaviors are selected. It is recommended in cascade generalization to begin with a low variance algorithm like Na\"ive Bayes~\citep{Zhang04} to draw stable decision surfaces ($H_{0}$ in Figure~\ref{fig:LCE}) and then use a low bias algorithm like boosting~\citep{Freund96} to fit more complex ones ($H_{1}$ in Figure~\ref{fig:LCE}). LC applies cascade generalization locally following a divide-and-conquer strategy based on mixture of experts. The objective of this approach is to capture new relationships that cannot be discovered globally. The LC divide-and-conquer method is a decision tree. When growing the tree, new attributes (class probabilities from a base classifier) are computed at each decision node and propagated down the tree. In order to be applied as a predictor, local cascade stores, in each node, the model generated by the base classifier.

\textbf{Local Cascade Ensemble} The contribution of LCE intervenes in the explicit manner of handling the bias-variance trade-off. Whereas LC approach is implicit, alternating between base classifiers behaviors (bias reduction, variance reduction) at each level of the tree, LCE is a hybrid ensemble method which combines an explicit boosting-bagging approach to handle the bias-variance trade-off and, as LC, an implicit divide-and-conquer approach - a decision tree. Firstly, LCE reduces bias across decision tree divide-and-conquer approach through the use of boosting-based classifiers as base classifiers ($H_{b}$ in Figure~\ref{fig:LCE}). A boosting-based classifier iteratively changes the data distribution with its reweighting scheme which decreases the bias. We adopt the best performing state-of-the-art boosting algorithm (eXtreme Gradient Boosting - XGB~\citep{Chen16}) as base classifier. In addition, boosting is propagated down the tree by adding the class probabilities of the base classifier as new attributes to the dataset. Class probabilities indicate the ability of the base classifier to correctly classify a sample. At the next tree level, class probabilities added to the dataset are exploited by the base classifier as a weighting scheme to focus more on previously misclassified samples. Then, the overfitting generated by the boosted decision tree is mitigated by the use of bagging. Bagging provides variance reduction by creating multiple predictors from random sampling with replacement of the original dataset (see $D^{1}$\ldots$D^{n}$ in Figure~\ref{fig:LCE}). Trees are aggregated with a simple majority vote.

The hybrid ensemble method LCE allows to balance bias and variance while benefiting from the improved generalization ability of explicitly creating different training sets (bagging, boosting). Furthermore, LCE implicit divide-and-conquer method ensures that classifiers are learned on different parts of the training data.

\vspace{-1em}
\subsection{XEM}
\label{XEM_Presentation}
As previously introduced, MTS classification has received significant interests over the past decade driven by automatic and high-resolution monitoring applications. A subset of the MTS can be characteristic of the event we aim to predict and can be adequate for the prediction.
Thus, we propose to leverage LCE tabular classifier to identify the discriminative part of an MTS and form an eXplainable-by-design Ensemble method for MTS classification (XEM) combining both performance and faithful explainability.
We have chosen a tabular classifier as the classifier of the MTS subsequences needs to learn the relation between the variables and potentially handle small time windows (e.g., length of 2), which prevents the use of most MTS classifiers.
Plus, we have selected LCE as it outperforms the state-of-the-art tabular classifiers on the public UCI datasets (see section~\ref{Res_LCE}).
The time window size is set as a parameter of XEM, which gives the estimated size of the discriminative part of an MTS. In our evaluation, without having prior knowledge on the time window size which would suit the classification tasks, we set the time window size using grid search with cross-validation (see section~\ref{Hyperparameters}).
In the following sections, we first present how dividing the time series into time windows is used to help XEM classify MTS based on their discriminative part and then, how it provides explainability-by-design.

\subsubsection{MTS Dataset Transformation}
\label{Dataset_Transformation}
\begin{figure}[!htpb]
	\centering
	\includegraphics[width=.9\linewidth]{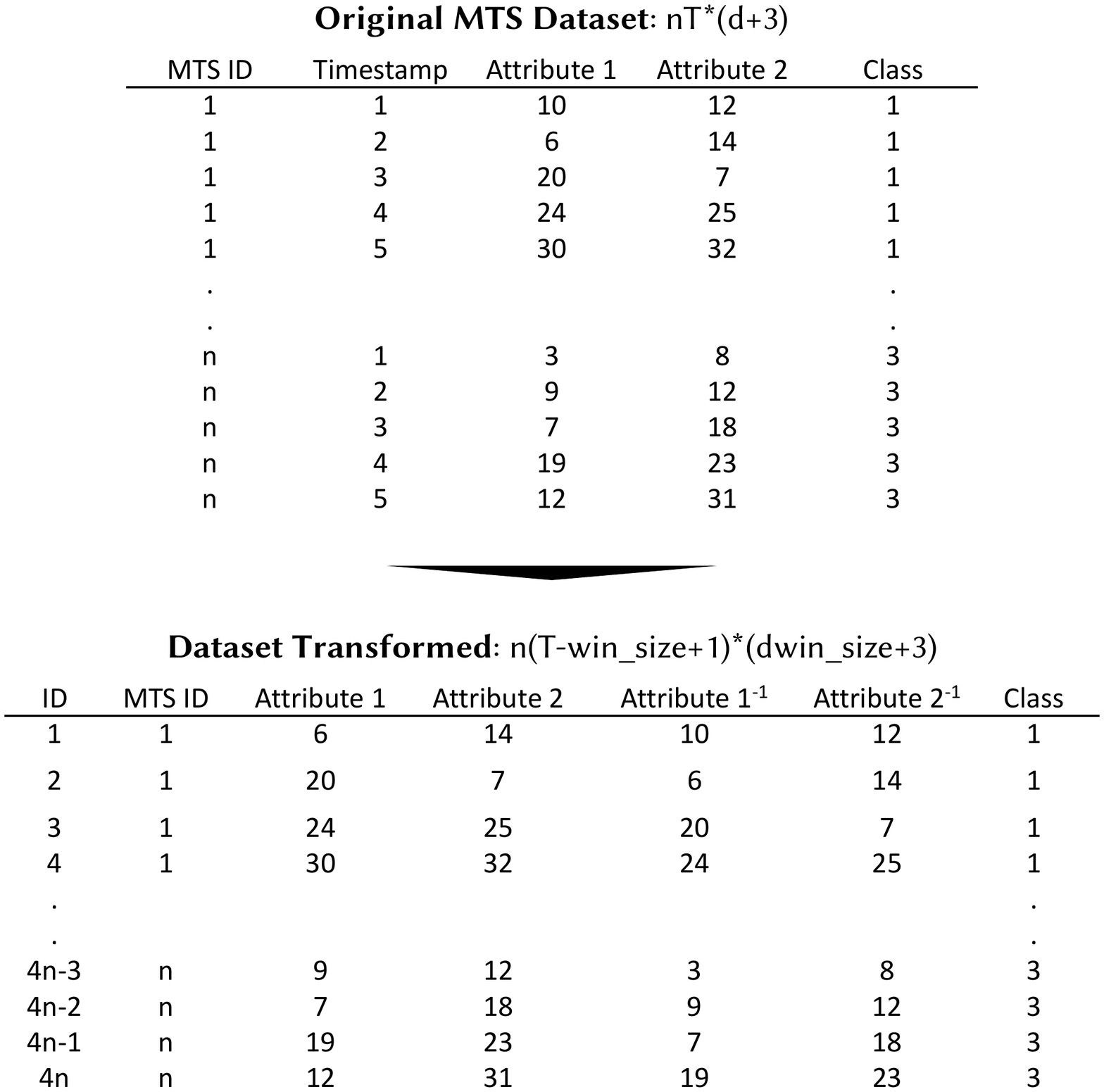}
	\caption{The dataset transformation (from original MTS to a flat dataset). AttributeX - value of attribute X, d - number of attributes, ID - sample identifier, MTS ID - MTS identifier, n - number of MTS, T - time series length, win\_size - time window size. In this example: T=5, d=2 and win\_size=2.}
	\label{fig:Dataset_Transformation}
\end{figure}

XEM trains LCE on subsequences of MTS to identify the discriminative time window, which requires a transformation of the original MTS dataset. This transformation is presented in Figure~\ref{fig:Dataset_Transformation}. Using a sliding window, all subsequences corresponding to the time window size are generated (MTS length $-$ time window size $+$ 1 subsequences). The time aspect is managed by setting the different timestamps as column dimensions. Each subsequence is considered as a new sample, labeled as the original MTS. For example in Figure~\ref{fig:Dataset_Transformation}, 4 subsequences (samples) are generated from the first MTS, composed of 2 timestamps (time window size) with 2 dimensions each (4 attributes columns). The 4 subsequences are calculated as: 5 (MTS length) $-$ 2 (time window size) $+$ 1. We present in the next sections how we compute the classification performance with the transformed dataset and how this configuration allows explainability.

\vspace{-1em}
\vspace{-2em}

\subsubsection{Classification}
\label{Algo_Classification}
\begin{figure}[!htpb]
	\centering
	\includegraphics[width=\linewidth]{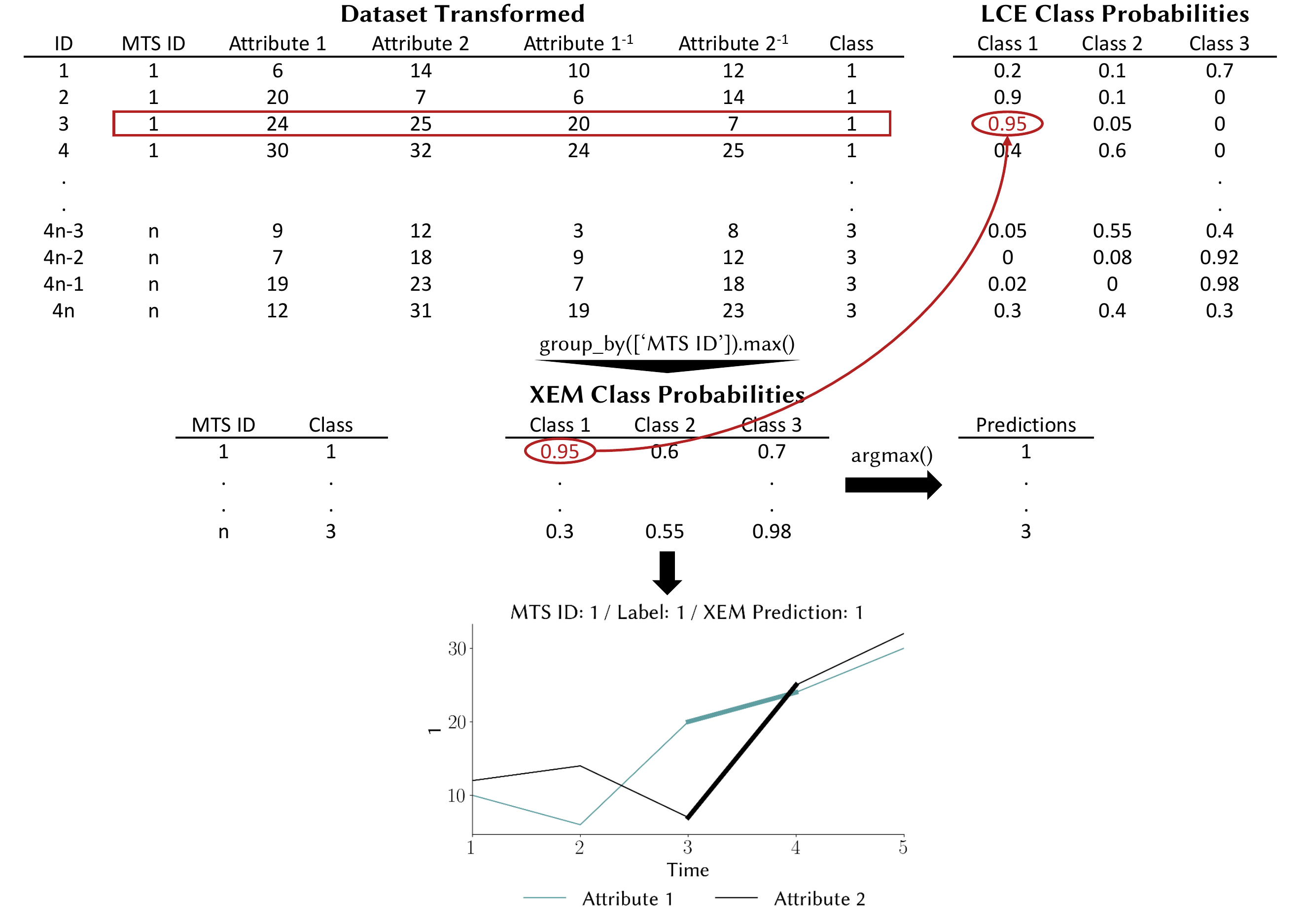}
	\caption{XEM prediction computation on the example from Figure~\ref{fig:Dataset_Transformation} with the identification of the discriminative time window for the MTS 1. And, an illustration of the explanation provided to the end-user to support XEM prediction for this MTS (highlighted in bold).}
	\label{fig:Explainability}
\end{figure}

As seen in the previous section, XEM trains LCE on samples corresponding to subsequences of MTS which sizes are controlled by the time window size parameter. 
Then, XEM assigns LCE class probabilities to all subsequences of the MTS. 
For example, on the upper part of Figure~\ref{fig:Explainability}, XEM assigns LCE class probabilities for each of the 4 subsequences of an MTS. Finally, XEM determines the class of an MTS based on the subsequence on which LCE is the most confident. For each MTS, the maximum class probability over the different subsequences is selected to determine the whole MTS classification output. For example, on the middle part of Figure~\ref{fig:Explainability}, we can observe that XEM assigns the class 1 to the first MTS (MTS ID=1) based on the highest class probability (0.95 versus 0.6 and 0.7) obtained with the classification of the third subsequence of the MTS. In the case where XEM is the most confident for a subsequence of an MTS which is not discriminative, it means that the time window size value is not suited for the classification problem and it would lead to poor classification accuracy of XEM on the training set. A time window size better suited for the classification problem would lead to better accuracy on the training set and would therefore be selected. The transformation presented and the performance evaluation procedure allow any traditional (tabular) classifier to perform MTS classification. Therefore, we compare in section~\ref{XEM_Performance} the performance of XEM to the best two state-of-the-art tabular classifiers applying the same transformation as LCE and to the state-of-the-art MTS classifiers.

\vspace{-2em}

\subsubsection{Explainability}
XEM prediction for an MTS is based on the subsequence that has the highest class probability - the subsequence on which LCE is the most confident. Thus, XEM provides explainability-by-design through the identification of the time window used to classify the MTS.
We illustrate the explainability of XEM with the previous section example in the lower part of Figure~\ref{fig:Explainability}. We observe that for the first MTS (MTS ID=1), after performing a grouping by MTS ID and taking the maximum, class 1 has the highest probability (0.95). We can trace back to the subsequence from which XEM is predicting this class probability (third subsequence), and show it to the end-user. This subsequence can help the end-user to understand why the MTS classifier attributed a particular label to the whole MTS (explainability). 
In this case, the subsequence associated with XEM prediction of the first MTS contains a steep increase of attribute 2 (black line - Figure~\ref{fig:Explainability}), which surpasses attribute 1 (blue line - Figure~\ref{fig:Explainability}).
We further illustrate the explainability property of XEM in section~\ref{Res_Explainability} on a synthetic and two UEA datasets.

\vspace{-1em}
\subsection{Properties}
In addition to its explainability-by-design, XEM has other interesting properties: phase invariance, interplay of dimensions, different MTS length compatibility, missing data management, noise robustness and scalability.

\begin{itemize}
	\item[$\bullet$] \textit{Phase Invariance}: XEM is not sensitive to the position of the discriminative subsequence in the MTS due to the selection of the subsequence which has the highest class probability to classify the whole MTS. This property improves the generalization ability of the algorithm: in the possible cases when the sequences of events in an MTS change, the classification result is not modified. For example, the classification result would be the same if the discriminative subsequence appears at the beginning or at the end of the MTS;
	\item[$\bullet$] \textit{Interplay of Dimensions}: XEM exploits the relationships among the dimensions through the use of boosting-based classifier as base classifier. It allows XEM to exploit complex interactions among dimensions at different timestamps to perform classification;
	\item[$\bullet$] \textit{Different MTS Length Compatibility}: XEM handles it in two different ways. If an MTS length is inferior to the maximum length of the MTS in a dataset multiplied by the window size selected, XEM uses padding of 0 values. Otherwise, no padding is necessary, less samples are generated per MTS but the performance evaluation procedure presented in~\ref{Algo_Classification} remains valid;
	\item[$\bullet$] \textit{Missing Data Management}: XEM naturally handles missing data through its tree-based learning~\citep{Breiman84}. Similar to extreme gradient boosting~\citep{Chen16}, XEM excludes missing values for the split and uses block propagation. 
	During a node split, block propagation sends all samples with missing data to the side minimizing the error, i.e. the node (left or right) which gets the highest score (accuracy score in this paper).
	We evaluate this property in our experiments in section~\ref{Missing_Data};
	\item[$\bullet$] \textit{Noise Robustness}: the bagging component of XEM provides noise robustness through variance reduction by creating multiple predictors from random sampling with replacement of the original dataset. We discuss this property in our experiments in section~\ref{Noise};
	\item[$\bullet$] \textit{Scalability}: as a tree-based ensemble method, XEM is scalable. Its time complexity is detailed in section~\ref{Time_Complexity}.
\end{itemize}

Most of the properties of XEM are coming from LCE. The properties shared between LCE and XEM are interplay of dimensions, missing data management, noise robustness and scalability.

\vspace{-1em}
\subsection{Time Complexity}
\label{Time_Complexity}
XEM time complexity corresponds to LCE time complexity plus the dataset transformation which is linear in the number of samples.
LCE time complexity is determined by the time complexity of multiple decision trees learning and extreme gradient boosting. The time complexity of building a single tree is $O(ndD_t)$, where $n$ is the number of samples, $d$ is the number of dimensions and $D_t$ is the maximum depth of the tree. So the time complexity of creating multiple decision trees with bagging is $O(N_tndD_t)$, where $N_t$ is the number of trees. Extreme gradient boosting has a time complexity of $O(N_bD_b\| x\|_0\log(n))$ where $N_b$ is the number of trees, $D_b$ is the maximum depth of the trees and $\| x\|_0$ is the number of non-missing entries in the data. Therefore, LCE has a time complexity of $O(N_tndD_t2^{D_t}N_bD_b\| x\|_0\log(n))$, where $2^{D_t}$ represents the maximum number of nodes in a binary tree. 
Table~\ref{tab:time_complexities} shows the time complexity of LCE in comparison with RF, XGB and LC.

\vspace{-1em}
\begin{table}[!htpb]
	\caption{Time complexities of the ensemble methods. $d$ - number of dimensions, $d'$ - number of dimensions in RF subset of dimensions, $D$ - maximum depth of a tree, $n$ - number of samples, $N$ - number of trees, $T_{Base}$ - time complexity of a base classifier, $\| x\|_0$ - number of non-missing entries in the data.}
	\scriptsize
	\centering
	\label{tab:time_complexities}
	\begin{tabularx}{.7\columnwidth}{>{\centering}m{3cm}>{\centering\arraybackslash}m{3.7cm}}
		\toprule
		\textbf{Algorithm} & \textbf{Time Complexity}\\
		\midrule\midrule[.1em]
		\textbf{RF} & $O(Nnd'D)$\\
		\textbf{XGB} & $O(N\log(n)\| x\|_0D)$\\
		\textbf{LC} & $O(ndD2^{D}T_{Base})$\\
		\textbf{LCE} & $O(NndD2^{D}T_{Base})$\\
		\bottomrule
	\end{tabularx}
\end{table} 
\vspace{-3em}

\subsection{Implementation}
\vspace{-2em}
\begin{algorithm}[!htpb]
	\caption{XEM}
	\label{xem}
	\begin{algorithmic}[1]
		\Require{A dataset $D$, a set of classifiers $H$, time window size $win\_size$, maximum depth of a tree $max\_depth$, number of trees $n\_trees$}
		
		\Function{XEM}{$D$, $H$, $win\_size$, $n\_trees$, $max\_depth$}
		\State $D' \leftarrow$ Dataset\_Transformation($D$, $win\_size$)
		\State $F \leftarrow \emptyset$
		\For{each $i$ in $\left[1, n\_trees\right]$}
		\State $S \leftarrow$ A bootstrap sample from $D'$
		\State $t \leftarrow$ XEM\_Tree($S$, $H$, $max\_depth$, $0$)
		\State $F \leftarrow F \cup t$
		\EndFor
		\State \Return $F$
		\EndFunction
		
		\Function{XEM\_Tree}{$D$, $H$, $max\_depth$, $depth$}
		\If{$max\_depth$ or uniform class}
		\State \Return leaf
		\Else
		\State $D^{\prime} \leftarrow$ Concatenate($D, H_{depth}(D)$)
		\State Split $D^{\prime}$ on attribute maximizing Gini criterion
		\State $depth \leftarrow depth + 1$
		\For{$D^{\prime(j)} \in \mathcal{P}(D^{\prime})$}
		\State $Tree_j =$ XEM\_Tree($D^{\prime(j)}$, $H$, $max\_depth$, $depth$)
		\EndFor
		\State \multiline{\Return tree containing one decision node, storing classifier $H_{depth}(D)$ and descendant subtrees $Tree_j$}
		\EndIf
		\EndFunction
	\end{algorithmic}	
\end{algorithm}
\vspace{-1em}

We present XEM pseudocode in Algorithm~\ref{xem} and make available our implementation\footnote{\label{xem_implementation}\url{https://github.com/XAIseries/XEM}} in Python 3.6. A function (XEM\_Tree) builds a tree and the second one (XEM) builds the forest of trees through bagging, after having transformed the dataset. There are 2 stopping criteria during a tree building phase: when a node has an unique class or when the tree reaches the maximum depth. We set the range of tree depth from 0 to 2 in XEM as in LCE. This hyperparameter is used to control overfitting. Low bias boosting-based classifier as base classifier justifies the maximum depth of 2. The set of low bias base boosting-based classifiers is limited to the best performing state-of-the-art boosting algorithm (XGB~\citep{Chen16}).

\vspace{-1em}
\section{Evaluation}
\label{Evaluation}
In this section, we present the methodology employed (datasets, algorithms, hyperparameters and metrics) to evaluate LCE and XEM.

\vspace{-1em}
\begin{table*}[!htpb]
	\caption{UCI datasets. Dims - Dimensions.}
	\label{tab:Datasets_Tabular}
	\centering
	\scriptsize
	\begin{tabularx}{\linewidth}{l>{\centering}m{1.5cm}>{\centering}m{1cm}>{\centering}m{1cm}|>{\centering}m{1cm}>{\centering\arraybackslash}m{1cm}}
		\toprule
		\multirow{2}{*}{\textbf{Datasets}} & \multirow{2}{*}{\textbf{Instances}} & \multirow{2}{*}{\textbf{Dims}} & \multirow{2}{*}{\textbf{Classes}} & \multicolumn{2}{c}{\textbf{LCE Parameters}} \\
		& & & & \textbf{Trees} & \textbf{Depth}\\
		\midrule\midrule[.1em]
		Absenteeism at Work & 740 & 19 & 19 & 100 & 2\\
		Banknote Authentification & 1372 & 4 & 2 & 5 & 1\\
		Breast Cancer Coimbra & 116 & 9 & 2 & 60 & 0\\
		CNAE-9 & 1,080 & 856 & 9 & 20 & 2\\
		Congressional Voting & 435 & 16 & 2 & 1 & 1\\
		Drug Consumption (quantified) & 1,185 & 12 & 7 & 5 & 2\\
		Electrical Grid Stability & 10,000 & 13 & 2 & 40 & 1\\
		Gas Sensor & 58 & 432 & 4 & 100 & 0\\
		HTRU2 & 17,898 & 8 & 2 & 60 & 2\\
		Iris & 150 & 4 & 3 & 20 & 2\\
		Leaf & 340 & 13 & 30 & 5 & 0\\
		LSVT Voice Rehabilitation & 126 & 310 & 2 & 5 & 0\\
		Lung Cancer & 32 & 56 & 3 & 60 & 1\\
		Mice Protein Expression & 1,080 & 77 & 8 & 60 & 1\\
		Musk V1 & 476 & 166 & 2 & 5 & 2\\
		Musk V2 & 6,598 & 166 & 2 & 5 & 2\\
		p53 Mutants & 31,159 & 5,408 & 2 & 10 & 1\\
		Page Blocks Classification & 5473 & 10 & 5 & 80 & 2\\
		Parkinson Disease & 756 & 753 & 2 & 5 & 2\\
		Semeion Handwritten Digit & 1,593 & 256 & 10 & 20 & 2\\
		Ultrasonic Flowmeter & 181 & 43 & 4 & 60 & 1\\
		User Knowledge Modeling & 403 & 5 & 5 & 40 & 2\\
		Wholesale Customers & 440 & 6 & 2 & 40 & 0\\
		Wine & 178 & 13 & 3 & 100 & 0\\
		Wine Quality & 1,599 & 11 & 6 & 100 & 2\\
		Yeast & 1,484 & 8 & 10 & 80 & 2\\
		\bottomrule
	\end{tabularx}
\end{table*}

\vspace{-2em}
\subsection{Datasets}
\vspace{-.5em}
\subsubsection{Multivariate Data}
\label{Dataset_Tabular}
In the experiments, we benchmarked LCE on the UCI datasets~\citep{Dua17}. We randomly selected one dataset per category available on the repository and obtained 26 UCI datasets. The categories are defined according to the number of instances (less than 100, 100 to 1,000 and greater than 1,000) and the number of dimensions (less than 10, 10 to 100 and greater than 100). The characteristics of each dataset are presented in Table~\ref{tab:Datasets_Tabular}. There is no train/test split provided on the repository so we decided to perform a stratified 3-fold cross-validation. 
Table~\ref{tab:Datasets_Tabular} also shows the values of LCE hyperparameters ($n\_trees$, $max\_depth$) set by grid search for each dataset during our experiments (see section~\ref{Hyperparameters} for hyperparameters optimization).

\subsubsection{Multivariate Time Series}
We benchmarked XEM on the 30 currently available UEA MTS datasets~\citep{Bagnall18}. We kept the train/test splits provided in the archive. The characteristics of each dataset are presented in Table~\ref{tab:Datasets}. 
Table~\ref{tab:Datasets} also shows the values of XEM hyperparameters ($n\_trees$, $max\_depth$, $win\_size$) set by grid search for each dataset during our experiments (see section~\ref{Hyperparameters} for hyperparameters optimization).

\vspace{-1em}
\begin{table*}[!htpb]
	\caption{UEA MTS datasets.\\ AS - Audio Spectra, C - Number of classes, De - Depth, Di - Dimensions, ECG - Electrocardiogram, EEG - Electroencephalogram, HAR - Human Activity Recognition, L - Time Series Length, MEG - Magnetoencephalography, Parameters - XEM Parameters, T - Number of trees, W - Time Window (\%).}
	\label{tab:Datasets}
	\centering
	\scriptsize
	\begin{tabularx}{\linewidth}{l>{\centering}m{1.1cm}>{\centering}m{.6cm}>{\centering}m{.6cm}>{\centering}m{.5cm}>{\centering}m{.3cm}>{\centering}m{.25cm}|>{\centering}m{.3cm}>{\centering}m{.2cm}>{\centering\arraybackslash}m{.2cm}}
		\toprule
		\multirow{2}{*}{\textbf{Datasets}} & \multirow{2}{*}{\textbf{Type}} & \multirow{2}{*}{\textbf{Train}} & \multirow{2}{*}{\textbf{Test}} & \multirow{2}{*}{\textbf{L}} & \multirow{2}{*}{\textbf{Di}} & \multirow{2}{*}{\textbf{C}} & \multicolumn{3}{c}{\textbf{Parameters}} \\
		& & & & & & & \textbf{W} & \textbf{T} & \textbf{De}\\
		\midrule\midrule[.1em]
		Articulary Word Recognition & Motion & 275 & 300 & 144 & 9 & 25 & 40 & 5 & 1\\
		Atrial Fibrilation & ECG & 15 & 15 & 640 & 2 & 3 & 20 & 1 & 0\\
		Basic Motions & HAR & 40 & 40 & 100 & 6 & 4 & 20 & 1 & 0\\
		Character Trajectories & Motion & 1,422 & 1,436 & 182 & 3 & 20 & 80 & 10 & 2\\
		Cricket & HAR & 108 & 72 & 1,197 & 6 & 12 & 40 & 20 & 0\\
		Duck Duck Geese & AS & 60 & 40 & 270 & 1,345 & 5 & 100 & 20 & 0\\
		Eigen Worms & Motion & 128 & 131 & 17,984 & 6 & 5 & 100 & 20 & 1\\
		Epilepsy & HAR & 137 & 138 & 206 & 3 & 4 & 20 & 1 & 1\\
		Ering & HAR & 30 & 30 & 65 & 4 & 6 & 20 & 1 & 2\\
		Ethanol Concentration & Other & 261 & 263 & 1751 & 3 & 4 & 20 & 1 & 2\\
		Face Detection & EEG/MEG & 5,890 & 3,524 & 62 & 144 & 2 & 100 & 5 & 2\\
		Finger Movements & EEG/MEG & 316 & 100 & 50 & 28 & 2 & 60 & 5 & 2\\
		Hand Movement Direction & EEG/MEG & 320 & 147 & 400 & 10 & 4 & 80 & 20 & 2\\
		Handwriting & HAR & 150	& 850 & 152 & 3 & 26 & 20 & 10 & 2\\ 
		Heartbeat & AS & 204 & 205 & 405 & 61 & 2 & 80 & 10 & 0\\
		Insect Wingbeat & AS & 30,000 & 20,000 & 200 & 30 & 10 & 100 & 10 & 1\\
		Japanese Vowels & AS & 270 & 370 & 29 & 12 & 9 & 40 & 5 & 1\\
		Libras & HAR & 180 & 180 & 45 & 2 & 15 & 40 & 60 & 1\\
		LSST & Other & 2,459 & 2,466 & 36 & 6 & 14 & 60 & 10 & 2\\
		Motor Imagery & EEG/MEG & 278 & 100 & 3,000 & 64 & 2 & 100 & 20 & 1\\
		NATOPS & HAR & 180 & 180 & 51 & 24 & 6 & 40 & 10 & 0\\
		PenDigits & Motion & 7,494 & 3,498 & 8 & 2 & 10 & 80 & 80 & 2\\
		PEMS-SF & Other & 267 & 173 & 144 & 963 & 7 & 100 & 20 & 1\\
		Phoneme & AS & 3315 & 3353 & 217 & 11 & 39 & 80 & 1 & 2\\
		Racket Sports & HAR & 151 & 152 & 30 & 6 & 4 & 60 & 20 & 0\\
		Self Regulation SCP1 & EEG/MEG & 268 & 293 & 896 & 6 & 2 & 100 & 5 & 2\\
		Self Regulation SCP2 & EEG/MEG & 200 & 180 & 1152 & 7 & 2 & 100 & 20 & 2\\
		Spoken Arabic Digits & AS & 6,599 & 2,199 & 93 & 13 & 10 & 80 & 10 & 1\\
		Stand Walk Jump & ECG & 12 & 15 & 2,500 & 4 & 3 & 20 & 1 & 1\\
		U Wave Gesture Library & HAR & 120 & 320 & 315 & 3 & 8 & 60 & 1 & 0\\
		\bottomrule
	\end{tabularx}
\end{table*}

\vspace{-1em}
\subsection{Algorithms}
\label{Classifiers}

\vspace{-.5em}
\subsubsection{Classifiers}
\label{Classifiers_tabular}
As presented in section~\ref{RW_Classification}, we evaluate the performance of LCE in comparison to:

\begin{itemize}	
	\item[$\bullet$] Bagging and Boosting - BB (ensemble method - explicit): we implemented the algorithm based on the description of the paper~\citep{Kotsiantis05} with 25 sub-classifiers for both bagging and boosting. We used the BaggingClassifier\footnote{sklearn.ensemble.BaggingClassifier} with the DecisionTreeClassifier\footnote{\label{cart_implementation}sklearn.tree.DecisionTreeClassifier} and the AdaBoostClassifier\footnote{\label{ada_implementation}sklearn.ensemble.AdaBoostClassifier} public implementations~\citep{scikit-learn};
	
	\item[$\bullet$] Boost-Wise Pre-Loaded Mixture of Experts - BP (ensemble method - explicit boosting + implicit): we implemented the algorithm based on the description of the paper~\citep{Ebrahimpour12}, with one hidden layer per MLP expert and the recommended learning rates (experts: 0.1, gating network: 0.05). We used the AdaBoostClassifier\textsuperscript{\ref{ada_implementation}}~\citep{scikit-learn} and the Keras\textsuperscript{\ref{keras_implementation}} public implementations;

	\item[$\bullet$] Elastic Net - EN (regularized logistic regression): the logistic regression combining L1 and L2 regularization methods. We used the SGDClassifier\footnote{\label{en_implementation}sklearn.linear\_model.SGDClassifier} public implementation~\citep{scikit-learn};

	\item[$\bullet$] Local Cascade - LC (ensemble method - implicit): we implemented the algorithm based on the description of the paper~\citep{Gama00} (hyperparameter: maximum depth of the tree $[0,5]$). The low bias base classifier is set to XGB\textsuperscript{\ref{xgb_implementation}} and the low variance base classifier to Na\"ive Bayes\footnote{\label{nb_implementation}sklearn.naive\_bayes.GaussianNB}~\citep{scikit-learn};

	\item[$\bullet$] Local Cascade Ensemble - LCE (ensemble method - hybrid): the algorithm has been implemented in Python 3.6\textsuperscript{\ref{xem_implementation}} (hyperparameters: $n\_trees$ $\{1,5,10,20,40,60,\\80,100\}$, $max\_depth$ $\{0,1,2\}$). The base classifier is set to XGB\textsuperscript{\ref{xgb_implementation}};
	
	\item[$\bullet$] Multilayer Perceptron - MLP (neural network): we consider small MLPs due to the limited size of the datasets and the absence of pretrained networks. We used the implementation available in the package Keras\footnote{\label{keras_implementation}\url{https://keras.io/}} and limit the neural network architecture to 3 layers;
	
	\item[$\bullet$] Random Forest - RF (ensemble method - explicit): we used the RandomForestClassifier\footnote{\label{rf_implementation}sklearn.ensemble.RandomForestClassifier} public implementation~\citep{scikit-learn};
	
	\item[$\bullet$] Simple Ensemble Method - SE: we used the DecisionTreeClassifier\textsuperscript{\ref{cart_implementation}}, GaussianNB\textsuperscript{\ref{nb_implementation}} and SGDClassifier\textsuperscript{\ref{en_implementation}} public implementations~\citep{scikit-learn};
	
	\item[$\bullet$] Support Vector Machine - SVM: we used the SVC\footnote{sklearn.svm.SVC} public implementation~\citep{scikit-learn};
	
	\item[$\bullet$] Extreme Gradient Boosting - XGB (ensemble method - explicit): we used the implementation available in the xgboost package for Python\footnote{\label{xgb_implementation}\url{https://xgboost.readthedocs.io/en/latest/python/}}.
\end{itemize}

\vspace{-.5em}
\subsubsection{MTS Classifiers}
We compare our algorithm XEM to the best two tabular classifiers from the previous evaluation applying the same transformation as LCE and to the state-of-the-art MTS classifiers.
\begin{itemize}	
	\item[$\bullet$] DTW$_{D}$, DTW$_{I}$ and ED - with and without normalization (n): we report the results published in the UEA archive~\citep{Bagnall18};
	
	\item[$\bullet$] MLSTM-FCN: we used the implementation available\footnote{\url{https://github.com/houshd/MLSTM-FCN}} and ran it with the parameter settings recommended by the authors in the paper~\citep{Karim19} (128-256-128 filters, kernel sizes 8/5/3, initialization of convolution kernels Uniform He, reduction ratio of 16, 250 training epochs, dropout of 0.8, Adam optimizer) and with the following hyperparameters: batch size $\{8,64,128\}$, number of LSTM cells $\{8,64,128\}$;
	
	\item[$\bullet$] RFM: Random Forest for Multivariate time series classification. We used the public implementation\textsuperscript{\ref{rf_implementation}} with the transformation presented in section~\ref{Dataset_Transformation};
	
	\item[$\bullet$] WEASEL+MUSE: we used the implementation available\footnote{\url{https://github.com/patrickzib/SFA}} and ran it with the parameter settings recommended by the authors in the paper~\citep{Schafer17} (chi=2, bias=1, p=0.1, c=5 and L2R\_LR\_DUAL solver) and with the following hyperparameters: SFA word lengths $\{2,4,6\}$, SFA quantization method \{equi-depth, equi-frequency\}, windows length [4, max(MTS length)];
	
	\item[$\bullet$] XEM: the algorithm has been implemented in Python 3.6\textsuperscript{\ref{xem_implementation}} with the following hyperparameters: $n\_trees$ $\{1,5,10,20,40,60,80,100\}$, $max\_depth$ $\{0,1,2\}$, $win\_size$ $\{20\%,40\%,60\%,80\%,100\%\}$;
	
	\item[$\bullet$] XGBM: Extreme Gradient Boosting for Multivariate time series classification. We used the public implementation\textsuperscript{\ref{xgb_implementation}} with the transformation presented in section~\ref{Dataset_Transformation}.
\end{itemize}

\vspace{-.5em}
\subsection{Hyperparameters Optimization}
\label{Hyperparameters}
Classifiers and MTS classifiers hyperparameters have been set for each dataset based on a stratified 3-fold cross-validation on the training sets. More specifically, hyperparameters of LC, LCE, MLSTM-FCN and XEM have been set by grid search. WEASEL+MUSE hyperparameters are set by the solver L2R\_LR\_DUAL as recommended by the authors. Then, the hyperparameters of all the other classifiers (BB, BP, EN, MLP, SE, SVM, RF, XGB) are set by hyperopt, a sequential model-based optimization using a tree of Parzen estimators search algorithm~\citep{Bergstra11}. Hyperopt chooses the next hyperparameters decision from both the previous choices and a tree-based optimization algorithm. Tree of Parzen estimators meet or exceed grid search and random search performance for hyperparameters setting. We use the implementation available in the Python package hyperopt\footnote{\url{https://github.com/hyperopt/hyperopt}} and hyperas\footnote{\url{https://github.com/maxpumperla/hyperas}} wrapper for Keras.

\vspace{-.5em}
\subsection{Metrics}
For each dataset, we compute the classification accuracy. Then, we present the average rank and the number of wins/ties to compare the different classifiers on the same datasets. 
Finally, we present the critical difference diagram~\citep{Demsar06}, the statistical comparison of multiple classifiers on multiple datasets based on the nonparametric Friedman test, to show the overall performance of LCE and XEM. The diagram represents the average rank of the classifiers, and the classifiers whose performance are not significantly different (inferior to the critical difference) are linked by a bar. An example of critical difference diagram can be seen in Figure~\ref{fig:CDPlot_Tabular}.
We use the implementation available in R package scmamp\footnote{\url{https://www.rdocumentation.org/packages/scmamp/versions/0.2.55/topics/plotCD}}.

\section{Results}
\label{Results}
In this section, we begin by evaluating the performance of LCE compared to the state-of-the-art classifiers. Next, we compare the performance of XEM to the other MTS classifiers. Then, we show that the explainability of XEM can give insights to the end-user about XEM predictions. Finally, we assess the robustness of XEM (missing data, noise) and position it into the performance-explainability framework introduced in~\citep{Fauvel20Framework}.

\subsection{LCE}
\label{Res_LCE}
Table~\ref{tab:UCI} shows the classification results of the 10 classifiers on the 26 UCI datasets. The best accuracy for each dataset is denoted in boldface. We observe that the top 3 classifiers are ensemble methods: LCE obtains the best average rank (2.8), followed by RF in second position (rank: 3.0) and XGB in third position (rank: 3.3).

First of all, LCE obtains the best average rank with the first position on 35\% of the datasets (9 wins/ties). Based on the categorization of the UCI datasets presented in Table~\ref{tab:Datasets_Tabular}, we do not observe any influence of the number of instances, dimensions or classes on the performance of LCE relative to other classifiers. 

\begin{table*}[!htpb]
	\caption{Accuracy results on the UCI datasets. MP - MLP, SV - SVM, XB - XGB}
	\label{tab:UCI}
	\centering
	\scriptsize
	\begin{tabularx}{\linewidth}{l>{\centering}m{.35cm}>{\centering}m{.35cm}>{\centering}m{.35cm}>{\centering}m{.35cm}>{\centering}m{.35cm}>{\centering}m{.35cm}>{\centering}m{.35cm}>{\centering}m{.35cm}>{\centering}m{.35cm}>{\centering\arraybackslash}m{.35cm}}
		\toprule
		\textbf{Datasets} & \textbf{LCE} & \textbf{LC} & \textbf{XB} & \textbf{RF} & \textbf{SE} & \textbf{BB} & \textbf{BP} & \textbf{MP} & \textbf{SV} & \textbf{EN}\\
		\midrule\midrule[.1em]
		Absenteeism at Work & 42.7 & 27.6 &	\textbf{44.2} & 42.0 & 29.9 & 38.1 & 21.8 & 28.3 & 28.7 & 31.7\\
		Banknote Authentification & 99.3 & 98.9 & 99.6 & 99.1 & 97.7 & 99.1 & 98.9 & 89.5 & \textbf{100} & 98.8\\
		Breast Cancer Coimbra & \textbf{71.4} & 65.5 & 64.6 & 64.5 & 49.2 & 57.8 & 54.3 & 48.4 & 55.2 & 57.5\\
		CNAE-9 & 86.2 & 51.0 & 84.1 & 91.6 & 90.9 & 87.4 & 95.5 & \textbf{95.6} & 30.4 & 92.2\\
		Congressional Voting & \textbf{97.0} & 94.0 & 96.8 & 96.6 & 95.2 & 96.8 & 91.7 & 79.5 & 87.8 & 91.7\\
		Drug Consumption (quantified) & 34.6 & 27.9 & 37.8 & 38.5 & 27.3 & 37.2 & 40.2 & \textbf{40.3} & \textbf{40.3} & 39.3\\
		Electrical Grid Stability & \textbf{100} & 99.9 & \textbf{100} & \textbf{100} & 98.4 & 99.9 & 94.9 & 88.5 & 79.3 & 96.8\\
		Gas Sensor & 74.4 & 63.3 & 74.6 & \textbf{89.6} & 88.1 & 86.1 & 75.6 & 78.7 & 61.5 & 70.4\\
		HTRU2 & \textbf{97.9} & 97.8 & \textbf{97.9} & 97.8 & 97.8 & 97.8 & 97.5 & 96.8 & 91.1 & 97.6\\
		Iris & \textbf{96.7} & 90.2 & \textbf{96.7} & \textbf{96.7} & 96.1 & \textbf{96.7} & 75.4 & 44.4 & 95.4 & 83.0\\
		Leaf & 52.5 & 48.7 & 61.6 & \textbf{71.7} & 30.6 & 37.9 & 10.2 & 8.5 & 35.2 & 56.0\\
		LSVT Voice Rehabilitation & \textbf{81.0} & 57.1 & 77.0 & \textbf{81.0} & 69.0 & 76.9 & 66.7 & 66.7 & 66.7 & 66.7\\
		Lung Cancer & 41.1 & 47.2 & 34.4 & 37.2 & 45.6 & 45.6 & 46.1 & 37.2 & 36.7 & \textbf{52.8}\\
		Mice Protein Expression & \textbf{56.7} & 40.1 & 43.1 & 53.1 & 35.9 & 46.1 & 35.1 & 13.9 & 14.4 & 42.9\\
		Musk V1 & 73.3 & 63.5 & \textbf{76.1} & 72.5 & 70.6 & 75.2 & 66.8 & 57.4 & 56.5 & 72.3\\
		Musk V2 & 78.8 & 74.5 & 78.4 & 77.5 & 78.6 & 77.2 & 78.3 & 84.6 & \textbf{84.7} & 76.3\\
		p53 Mutants & 96.6 & 82.7 & 94.8 & 95.6 & 83.8 & 91.1 & 85.4 & \textbf{99.5} & 86.5 & 81.7\\	
		Page Blocks Classification & \textbf{97.3} & 90.8 & 96.5 & 96.0 & 94.2 & 95.4 & 93.6 & 90.4 & 91.1 & 94.2\\
		Parkinson Disease & 82.7 & 74.2 & 82.5 & \textbf{83.2} & 75.5 & 82.4 & 74.6 & 58.2 & 74.6 & 41.4\\
		Semeion Handwritten Digit & 90.3 & 43.2 & 90.0 & \textbf{92.2} & 77.3 & 83.9 & 90.8 & 92.1 & 36.4 & 75.8\\
		Ultrasonic Flowmeter & \textbf{59.0} & 40.2 & 45.2 & 49.6 & 42.8 & 48.4 & 36.9 & 24.4 & 29.8 & 45.1\\
		User Knowledge Modeling & 85.6 & 80.4 & 85.6 & 85.6 & 79.4 & \textbf{85.9} & 57.8 & 29.8 & 80.4 & 74.6\\
		Wholesale Customers & 91.8 & 88.6 & \textbf{92.5} & 91.6 & 85.2 & 91.6 & 76.3 & 77.0 & 67.7 & 83.0\\
		Wine & 92.8 & \textbf{96.1} & 91.1 & 92.8 & 89.4 & 87.6 & 39.9 & 35.4 & 42.7 & 75.4\\
		Wine Quality & 55.5 & 49.2 & 54.5 & \textbf{56.9} & 46.7 & 53.7 & 46.9 & 42.1 & 41.9 & 45.9\\
		Yeast & 57.1 & 35.3 & 59.2 & \textbf{59.6} & 47.6 & 57.5 & 34.1 & 28.9 & 58.9 & 53.2\\
		\hline
		Average Rank & 2.8 & 6.8 & 3.3 & 3.0 & 6.0 & 4.1 & 7.0 & 7.6 & 7.3 & 6.5 \\
		Wins/Ties & 9 & 1 & 6 & 9 & 0 & 2 & 0 & 3 & 3 & 1\\
		\bottomrule
	\end{tabularx}
\end{table*}

Then, we observe that the second ranked classifier RF obtains the same number of wins/ties as LCE (9 win/ties). 
RF gets around 60\% of its wins/ties on small datasets (train size $<$ 1000). We can infer that the bagging only (variance reduction) of RF can provide better generalization than LCE bagging-boosting combination on small datasets (wins/ties on small datasets - 54\% of the datasets: LCE 5, RF 5). The third ranked classifier XGB gets 6 wins/ties. We do not see any influence of the different dataset categories on XGB wins/ties relative to LCE. Therefore, we conclude that LCE bagging and boosting combination to handle the bias-variance trade-off exhibits better generalization on average than the bagging only (RF) and boosting only (XGB) algorithms on the UCI datasets.

Next, LC algorithm gets the fifth rank with one win/tie. We do not see any particular influence of the different dataset categories on LC performance. So, the outperformance of LCE compared to LC on the UCI datasets confirms the better generalization ability of a hybrid (explicit and implicit) versus an implicit only approach. The comparison in Table~\ref{tab:LCversusLCE} aims to underline the superior performance of LCE compared to LC on the UCI datasets. In order to be comparable, the depth of a tree is set to 1 for LC and LCE and, as presented in section~\ref{Classifiers_tabular}, the low bias base classifier in LC and LCE is the best performing state-of-the-art boosting algorithm - XGB. The results correspond to the average accuracy on test sets with the corresponding standard error. Results show a comparable accuracy variability of LCE compared to LC when the number of trees is set to 1 (standard error of 4.6\% versus 4.8\%). However, LCE on 1 tree exhibits a higher accuracy than LC (71.8\% versus 65.9\%). Additionally, through bagging, we observe LCE variability reduction as well as an increase of accuracy (71.8$\pm$4.6 with 1 tree versus 74.9$\pm$4.1 with 60 trees versus 65.9$\pm$4.8 with LC). Therefore, this comparison affirms the superiority of our explicit bias-variance trade-off approach compared to the implicit approach of LC on the UCI datasets.

\begin{table}[!htpb]
	\caption{Average accuracy score of LCE versus LC on test sets of the UCI datasets with the corresponding standard error.}
	\label{tab:LCversusLCE}
	\centering
	\scriptsize
	\begin{tabularx}{.8\columnwidth}{lYYYYYYY}
		\toprule
		\textbf{Trees} & \textbf{1} & \textbf{5} & \textbf{10} & \textbf{20} & \textbf{40} & \textbf{60} & \textbf{80}\\
		\midrule\midrule[.1em]
		\textbf{LCE} & 71.8 & 74.1 & 73.6 & 72.8 & 73.2 & 74.9 & 73.9\\
		& $\pm4.6$ & $\pm4.3$ & $\pm4.4$ & $\pm4.4$ & $\pm4.5$ & $\pm4.1$ & $\pm4.2$\\
		& & & & & & &\\
		\textbf{LC} & \multicolumn{7}{c}{$65.9\pm4.8$}\\
		\bottomrule
	\end{tabularx}
	\vspace{-1em}
\end{table}

Moreover, LCE hybrid approach shows better average performance than the remaining ensemble methods, and in particular the combination of explicit methods - BB, as well as the combination of the explicit boosting method with an implicit approach - BP (rank: LCE 2.8 , BB 4.1, SE 6.0, BP 7.0).
LCE outperforms BB, SE and BP on both small (rank: LCE 2.5, BB 3.4, SE 5.6, BP 7.6) and large datasets (rank: LCE 3.2, BB 4.9, SE 6.5, BP 6.2).

Concerning the other classifiers, EN obtains only one win/tie but gets a better rank on average than SVM (3 wins/ties) and MLP (3 wins/ties).

\begin{figure}[!htpb]
	\centering
	\includegraphics[width=.6\linewidth]{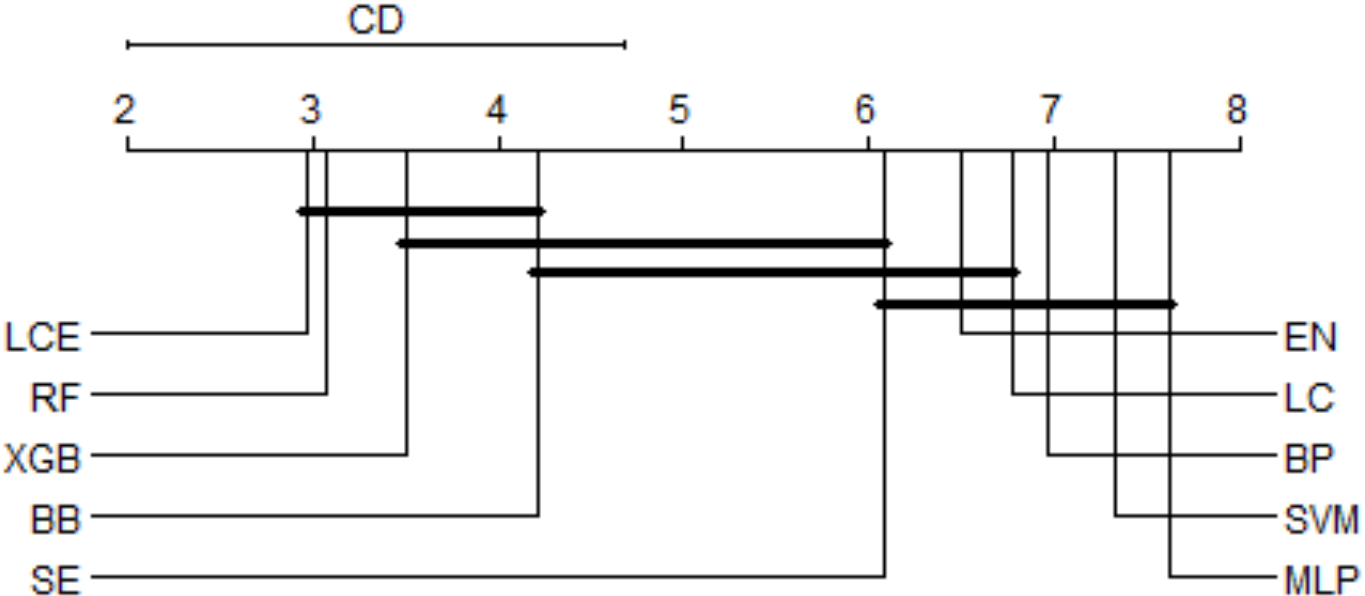}
	\caption{Critical difference plot of the classifiers on the UCI datasets with alpha equals to 0.05.}
	\label{fig:CDPlot_Tabular}
	\vspace{-1em}
\end{figure}

Finally, we analyze a statistical test to evaluate the performance of LCE compared to other classifiers. We present in Figure~\ref{fig:CDPlot_Tabular} the critical difference plot with alpha equals to 0.05 from results shown in Table~\ref{tab:UCI}. The values correspond to the average rank and the classifiers linked by a bar do not have a statistically significant difference. The plot confirms the top 3 ranking as presented before (LCE: 1, RF: 2, XGB: 3). 
We also observe that LCE and RF have a significant performance difference compared to SE.
Therefore, considering that LCE transformation to multivariate time series classification is also applicable to other traditional (tabular) classifiers, we evaluate the performance of RF and XGB with the same transformation as LCE in comparison to the state-of-the-art MTS classifiers in the next section.

\subsection{XEM}
\label{XEM_Evaluation}
\subsubsection{Classification Performance}
\label{XEM_Performance}
The classification results of the 11 MTS classifiers are presented in Table~\ref{tab:UEA}. A blank in the table indicates that the approach ran out of memory or the accuracy is not reported \citep{Bagnall18}. The best accuracy for each dataset is denoted in boldface. We observe that XEM obtains the best average rank (3.0), followed by RFM in second position (rank: 3.7) and MLSTM-FCN in third position (rank: 3.8).

\begin{table*}[!htpb]
	\caption{Accuracy results on the UEA MTS datasets.\\ D$_{D}$ - DTW$_{D}$, D$_{I}$ - DTW$_{I}$, MF - MLSTM-FCN, RM - RFM, WM - WEASEL+MUSE, XG - XGBM, XM - XEM}
	\label{tab:UEA}
	\centering
	\scriptsize
	\begin{tabularx}{\linewidth}{l>{\centering}m{.28cm}>{\centering}m{.28cm}>{\centering}m{.28cm}>{\centering}m{.28cm}>{\centering}m{.28cm}>{\centering}m{.28cm}>{\centering}m{.28cm}>{\centering}m{.28cm}>{\centering}m{.28cm}>{\centering}m{.28cm}>{\centering\arraybackslash}m{.28cm}}	
		\toprule
		\textbf{Datasets} & \textbf{XM} & \textbf{XG} & \textbf{RM} & \textbf{MF} & \textbf{WM} & \textbf{ED} & \textbf{D$_{I}$} & \textbf{D$_{D}$} & \textbf{ED (n)} & \textbf{D$_{I}$ (n)} & \textbf{D$_{D}$ (n)}\\
		\midrule\midrule[.1em]
		Articulary Word Recognition & \textbf{99.3} & 99.0 & 99.0 & 98.6 & \textbf{99.3} & 97.0 & 98.0 & 98.7 & 97.0 & 98.0 & 98.7\\
		Atrial Fibrilation & \textbf{46.7} & 40.0 & 33.3 & 20.0 & 26.7 & 26.7 & 26.7 & 20.0 & 26.7 & 26.7 & 22.0\\
		Basic Motions & \textbf{100} & \textbf{100} & \textbf{100} & \textbf{100} & \textbf{100} & 67.5 & \textbf{100} & 97.5 & 67.6 & \textbf{100} & 97.5\\
		Character Trajectories & 97.9 & 98.3 & 98.5 & \textbf{99.3} & 99.0 & 96.4 & 96.9 & 99.0 & 96.4 & 96.9 & 98.9\\
		Cricket & 98.6 & 97.2 & 98.6 & 98.6 & 98.6 & 94.4 & 98.6 & \textbf{100} & 94.4 & 98.6 & \textbf{100}\\
		Duck Duck Geese & 37.5 & 40.0 & 40.0 & \textbf{67.5} & 57.5 & 27.5 & 55.0 & 60.0 & 27.5 & 55.0 & 60.0\\
		Eigen Worms & 52.7 & 55.0 & \textbf{100} & 80.9 & 89.0 & 55.0 & 60.3 & 61.8 & 54.9 &  & 61.8\\
		Epilepsy & 98.6 & 97.8 & 98.6 & 96.4 & \textbf{99.3} & 66.7 & 97.8 & 96.4 & 66.6 & 97.8 & 96.4\\
		Ering & \textbf{20.0} & 13.3 & 13.3 & 13.3 & 13.3 & 13.3 & 13.3 & 13.3 & 13.3 & 13.3 & 13.3\\
		Ethanol Concentration & 37.2 & 42.2 & \textbf{43.3} & 29.4 & 31.6 & 29.3 & 30.4 & 32.3 & 29.3 & 30.4 & 32.3\\
		Face Detection & 61.4 & \textbf{62.9} & 61.4 & 57.4 & 54.5 & 51.9 & 51.3 & 52.9 & 51.9 &  & 52.9\\
		Finger Movements & 59.0 & 53.0 & 56.0 & \textbf{61.0} & 54.0 & 55.0 & 52.0 & 53.0 & 55.0 & 52.0 & 53.0\\
		Hand Movement Direction & \textbf{64.9} & 54.1 & 50.0 & 37.8 & 37.8 & 27.9 & 30.6 & 23.1 & 27.8 & 30.6 & 23.1\\
		Handwriting & 28.7 & 26.7 & 26.7 & 54.9 & 53.1 & 37.1 & 50.9 & \textbf{60.7} & 20.0 & 31.6 & 28.6\\ 
		Heartbeat & 76.1 & 69.3 & \textbf{80.0} & 71.4 & 72.7 & 62.0 & 65.9 & 71.7 & 61.9 & 65.8 & 71.7\\
		Insect Wingbeat & 22.8 & \textbf{23.7} & 22.4 & 10.5 &  & 12.8 &  & 11.5 & 12.8 &  &\\
		Japanese Vowels & 97.8 & 96.8 & 97.0 & \textbf{99.2} & 97.8 & 92.4 & 95.9 & 94.9 & 92.4 & 95.9 & 94.9\\
		Libras & 77.2 & 76.7 & 78.3 & \textbf{92.2} & 89.4 & 83.3 & 89.4 & 87.2 & 83.3 & 89.4 & 87.0\\
		LSST & \textbf{65.2} & 63.3 & 61.2 & 64.6 & 62.8 & 45.6 & 57.5 & 55.1 & 45.6 & 57.5 & 55.1\\
		Motor Imagery & \textbf{60.0} & 46.0 & 55.0 & 53.0 & 50.0 & 51.0 & 39.0 & 50.0 & 51.0 &  & 50.0\\
		NATOPS & 91.6 & 90.0 & 91.1 & \textbf{96.7} & 88.3 & 85.0 & 85.0 & 88.3 & 85.0 & 85.0 & 88.3\\
		PenDigits & 97.7 & 95.1 & 95.1 & \textbf{99.0} & 96.9 & 97.3 & 93.9 & 97.7 & 97.3 & 93.9 & 97.7\\
		PEMS-SF & 94.2 & \textbf{98.3} & \textbf{98.3} & 69.9 &  & 70.5 & 73.4 & 71.1 & 70.5 & 73.4 & 71.1\\
		Phoneme & \textbf{28.8} & 18.7 & 22.2 & 27.5 & 19.0 & 10.4 & 15.1 & 15.1 & 10.4 & 15.1 & 15.1\\
		Racket Sports & \textbf{94.1} & 92.8 & 92.1 & 89.4 & 91.4 & 86.4 & 84.2 & 80.3 & 86.8 & 84.2 & 80.3\\
		Self Regulation SCP1 & 83.9 & 82.9 & 82.6 & \textbf{86.7} & 74.4 & 77.1 & 76.5 & 77.5 & 77.1 & 76.5 & 77.5\\
		Self Regulation SCP2 & \textbf{55.0} & 48.3 & 47.8 & 52.2 & 52.2 & 48.3 & 53.3 & 53.9 & 48.3 & 53.3 & 53.9\\
		Spoken Arabic Digits & 97.3 & 97.0 & 96.8 & \textbf{99.4} & 98.2 & 96.7 & 96.0 & 96.3 & 96.7 & 95.9 & 96.3\\
		Stand Walk Jump & 40.0 & 33.3 & \textbf{46.7} & \textbf{46.7} & 33.3 & 20.0 & 33.3 & 20.0 & 20.0 & 33.3 & 20.0\\
		U Wave Gesture Library & 89.7 & 89.4 & 90.0 & 86.3 & \textbf{90.3} & 88.1 & 86.9 & \textbf{90.3} & 88.1 & 86.8 & \textbf{90.3}\\
		\hline
		Average Rank & 3.0 & 4.8 & 3.7 & 3.8 & 4.1 & 7.6 & 6.3 & 5.3 & 7.9 & 6.7 & 5.7\\
		Wins/Ties & 10 & 4 & 6 & 11 & 4 & 0 & 1 & 3 & 0 & 1 & 2\\
		\bottomrule
	\end{tabularx}
\end{table*}

XEM gets the first position in one third of the datasets. Using the categorization of the datasets published in the archive website\footnote{\url{http://www.timeseriesclassification.com/dataset.php}}, we do not see any influence from the different train set sizes, MTS lengths, dimensions and number of classes on XEM performance relative to the other classifiers on the UEA datasets.
Nonetheless, XEM exhibits weaker performance on average on human activity recognition (rank: 3.6, 30\% of all datasets) and motion classification (rank: 5.0, 13\% of all datasets) datasets.

Then, we observe that the better generalization of LCE bagging-boosting combination compared to bagging only (RF) and boosting only (XGB) is also valid on the MTS datasets (average rank: XEM 3.0, RFM 3.7, XGBM 4.8). The adaptation of ensemble methods to the MTS datasets (see section~\ref{Dataset_Transformation}) is well performing: the three ensemble methods obtain the highest number of wins/ties (ensemble methods for MTS: 17 - 57\% of all datasets, MLSTM-FCN: 11 - 37\% of all datasets, WEASEL+MUSE: 4 - 13\% of all datasets). The 6 wins/ties of RFM are obtained on small datasets (train size $<$ 500). As seen in section~\ref{Res_LCE}, we can infer that the bagging only (variance reduction) of RFM can provide better generalization than XEM bagging-boosting combination on small datasets (wins/ties on small datasets - 77\% of the datasets: XEM 8, RFM 6). 
On the time window sizes used, we observe that the choice of XEM time window is a trade-off between its bagging and boosting components. 
XEM and XGBM use the same time window size on 70\% of the datasets. When the time window size is different, XEM obtains a better accuracy than XGBM on 90\% of the cases. Moreover, XEM employs the same time window size as RFM on half of the UEA datasets. On the other half of the datasets, RFM adopts a slightly bigger time window size than XEM. RFM uses a bigger time window in 75\% of the time with an average time window difference of 29\% between XEM and RFM. The different choice of XEM time window size leads to a better accuracy on 75\% of the cases compared to RFM. These observations prove that XEM bias-variance trade-off can refine the time window size of boosting only and bagging only to obtain a better generalization ability on average.

\begin{figure}[!htpb]
	\centering
	\includegraphics[width=.7\linewidth]{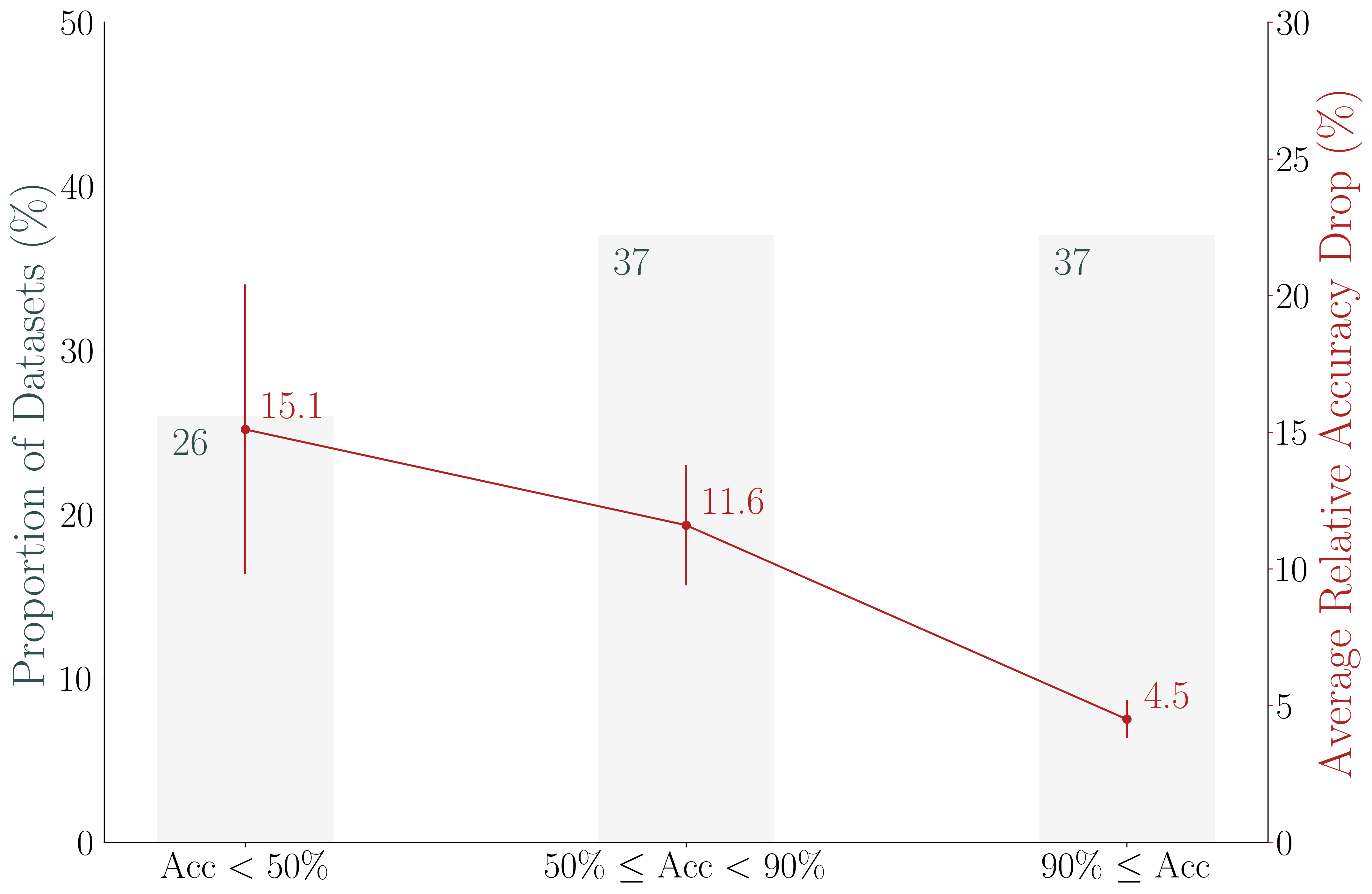}
	\caption{XEM average relative accuracy drop across the UEA datasets when using other time window sizes than the one used in the best configuration given in Table~\ref{tab:Datasets}. The performance drop is presented across three categories of datasets, defined according to XEM levels of accuracy shown in Table~\ref{tab:UEA}. Acc - Accuracy.}
	\label{fig:time_windows}
	\vspace{-1em}
\end{figure}

Specifically, with regard to the hyperparameter $win\_size$ of XEM, Figure~\ref{fig:time_windows} shows the average relative drop in performance across the datasets when using the other time window sizes than the one used in the best configuration given in Table~\ref{tab:Datasets}. 
In order to evaluate the relative impact with respect to the range of performance, we have defined three categories of datasets: datasets with XEM original accuracy $<$ 50\%, datasets with 50\% $\leq$ accuracy $<$ 90\% and datasets with accuracy $\geq$ 90\%.
First, as expected, we observe that the average relative impact of using suboptimal time window sizes is higher when XEM level of performance is low (average relative drop in accuracy: 15.1\% when XEM accuracy $<$ 50\% versus 4.5\% when XEM accuracy $\geq$ 90\%).
Then, the average relative drop in accuracy when using suboptimal time window sizes is not negligible but remains limited in all the cases. This drop is below 16\% on average on the category where XEM has the lowest level of accuracy (15.1\% $\pm$ 5.3\%) and below 10\% on average across all the datasets (9.9\% $\pm$ 1.8\%).

Concerning the state-of-the-art MTS classifiers, we observe a performance difference between the third (MLSTM-FCN) and fourth (WEASEL+MUSE) classifiers on datasets sizes. MLSTM-FCN outperforms WEASEL+MUSE (rank: 2.6 versus 4.6 for WEASEL+MUSE) on the largest datasets (train size $\geq$ 500, 23\% of all datasets) whereas WEASEL+MUSE slightly outperforms MLSTM-FCN (rank 4.0 versus 4.2 for MLSTM-FCN) on the smallest datasets (train size $<$ 500, 77\% of all datasets). XEM shows the same performance as MLSTM-FCN on the largest datasets (rank 2.6) while outperforming WEASEL+MUSE on the smallest datasets (rank: 3.2 versus 4.0 for WEASEL+MUSE). Therefore, XEM is better than the state-of-the-art MTS classifiers on both the small and large UEA datasets. Last, similarity-based methods obtain the lowest wins/ties counts. Euclidean distance is never in the first position on the UEA datasets. The wins/ties of DTW (DTW$_{D}$ normalized: 2, DTW$_{D}$: 3) stem from their outperformance on human activity recognition datasets.

\begin{figure}[!htpb]
	\centering
	\includegraphics[width=\linewidth]{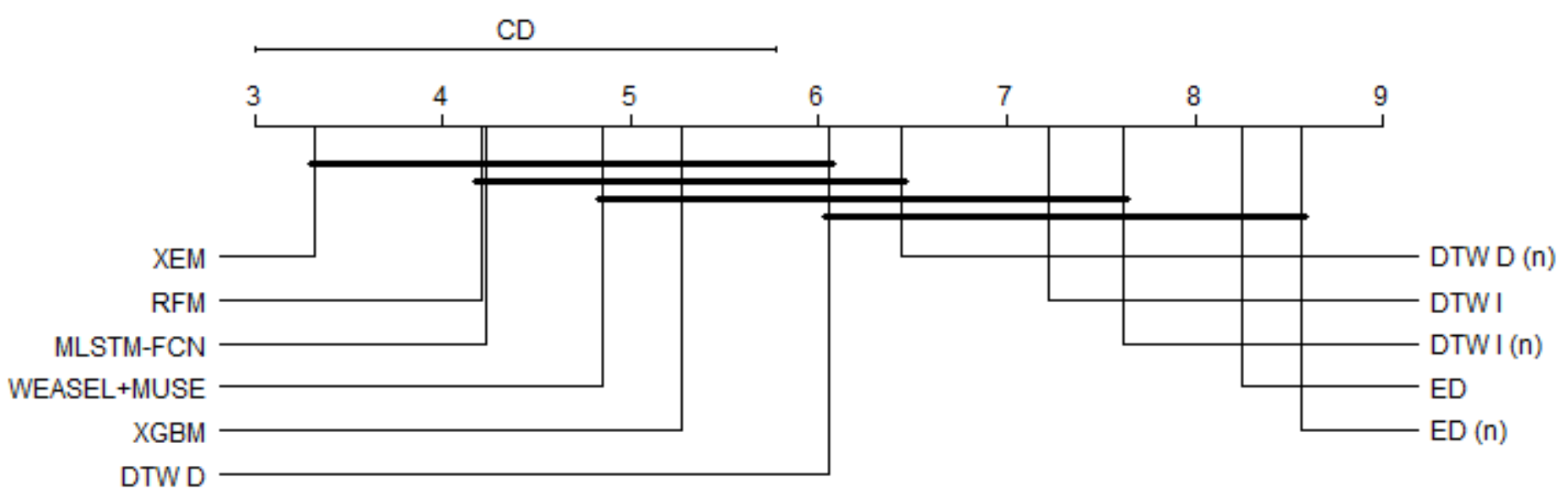}
	\caption{Critical difference plot of the MTS classifiers on the UEA datasets\\ with alpha equals to 0.05.}
	\label{fig:CDPlot}
	\vspace{-1em}
\end{figure}

Next, we performed a statistical test to evaluate the performance of XEM compared to other MTS classifiers. We present in Figure~\ref{fig:CDPlot} the critical difference plot with alpha equals to 0.05 from results shown in Table~\ref{tab:UEA}. The values correspond to the average rank and the classifiers linked by a bar do not have a statistically significant difference. The plot confirms the top 3 ranking as presented before (XEM: 1, RFM: 2, MLSTM-FCN: 3). We notice that XEM is the only classifier with a significant performance difference compared to DTW$_{D}$ normalized.

\subsubsection{XEM Explainability}
\label{Res_Explainability}
This section presents XEM explainability-by-design results. First, we illustrate the explainability of XEM on a synthetic dataset.
The construction of a synthetic dataset allows us to know the expected discriminative time window.
Then we show which windows have been used by XEM on the UEA datasets of section~\ref{XEM_Performance} and present the explainability results on two UEA datasets. 
We do not know the expected discriminative time windows on the UEA datasets so it is worth noting that the explanations provided on these two UEA datasets are given as illustrative in nature.
In addition, for each dataset, we compare XEM explainability-by-design results with the ones from certain post hoc model-agnostic explainability methods. The current best performing state-of-the-art MTS classifiers (MLSTM-FCN, WEASEL+MUSE) are black-box classifiers, which can only rely on post hoc model-agnostic explainability methods. Therefore, in order to emphasize the value coming from XEM explainability-by-design, we study the difference between XEM explainability results and the ones obtained from certain post hoc model-agnostic explainability methods applied to XEM. Multiple post hoc model-agnostic explainability methods exist (e.g., LIME~\citep{Ribeiro16}, SHAP~\citep{Lundberg17}, Anchors~\citep{Ribeiro18}, LORE~\citep{Guidotti19}, features tweaking~\citep{Karlsson20}). Among the post hoc model-agnostic explainability methods, we have chosen the type with feature importance as it is the most popular one, and similarly to XEM, it identifies the regions of the input data that are important for a particular prediction.	Specifically, we have chosen Local Interpretable Model-Agnostic Explanations (LIME) and SHapley Additive exPlanations (SHAP), the current state-of-the-art methods offering local explainability under the form of feature importance. These methods use an explainable surrogate model, a model that aims to mimic the predictions of the original one. More specifically, LIME describes the local behavior of the model using a linearly weighted combination of the input features, learned on perturbations of an instance. SHAP also adopts a linear surrogate model: an additive feature attribution method that uses simplified inputs (conditional expectations) assuming feature independence. Thus, these methods provide how much each variable (features+time) impacts predictions. We cannot apply LIME and SHAP methods at a higher granularity to obtain explanations at windows level (like XEM) as their surrogate models would combine information from multiple windows to mimic the performance of XEM, when XEM only uses one window to perform classification. For each dataset, in order to compare explainability results, we represent on the input data the identified regions that are important for predictions from XEM explainability-by-design, LIME and SHAP results.

\begin{figure*}[!htpb]
	\centering
	\includegraphics[width=\linewidth]{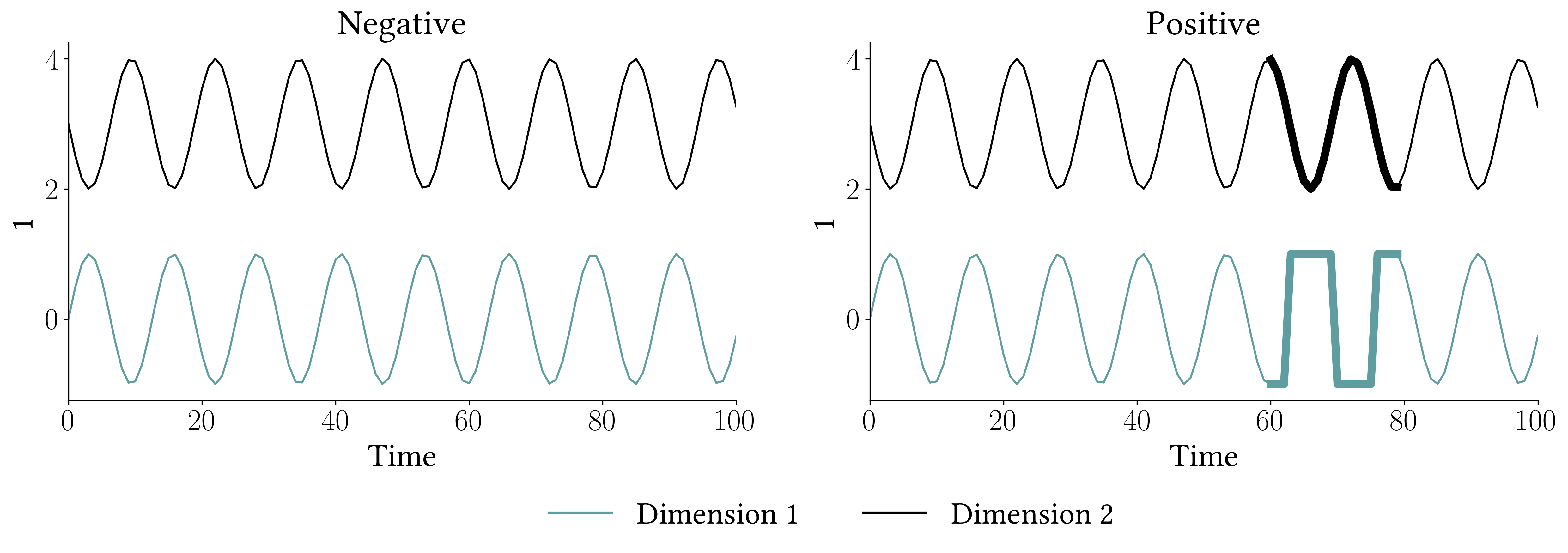}
	\caption{The two MTS types of the synthetic dataset, with the XEM time window used for the classification of MTS belonging to the \textit{positive} class highlighted in bold, which serves as the explanation for the end-user ($win\_size$: 20\%).}
	\label{fig:synthetic_intepretability}
	\vspace{-1em}
\end{figure*}

\textbf{Synthetic Dataset} First of all, we show that XEM uses and identifies the expected time window to perform the classification on an MTS synthetic dataset. We design a dataset composed of 20 MTS (50\%/50\% train/test split) with a length of 100, 2 dimensions and 2 balanced classes. The difference between the 10 MTS belonging to the \textit{negative} class and the one belonging to the \textit{positive} class stems from a 20\% time window of the MTS. As illustrated in Figure~\ref{fig:synthetic_intepretability}, \textit{negative} class MTS are sine waves and \textit{positive} class MTS are sine waves with a square signal on 20\% of the dimension 1 (see timestamps between 60 and 80).

\begin{figure}[!htpb]
	\centering
	\includegraphics[width=\linewidth]{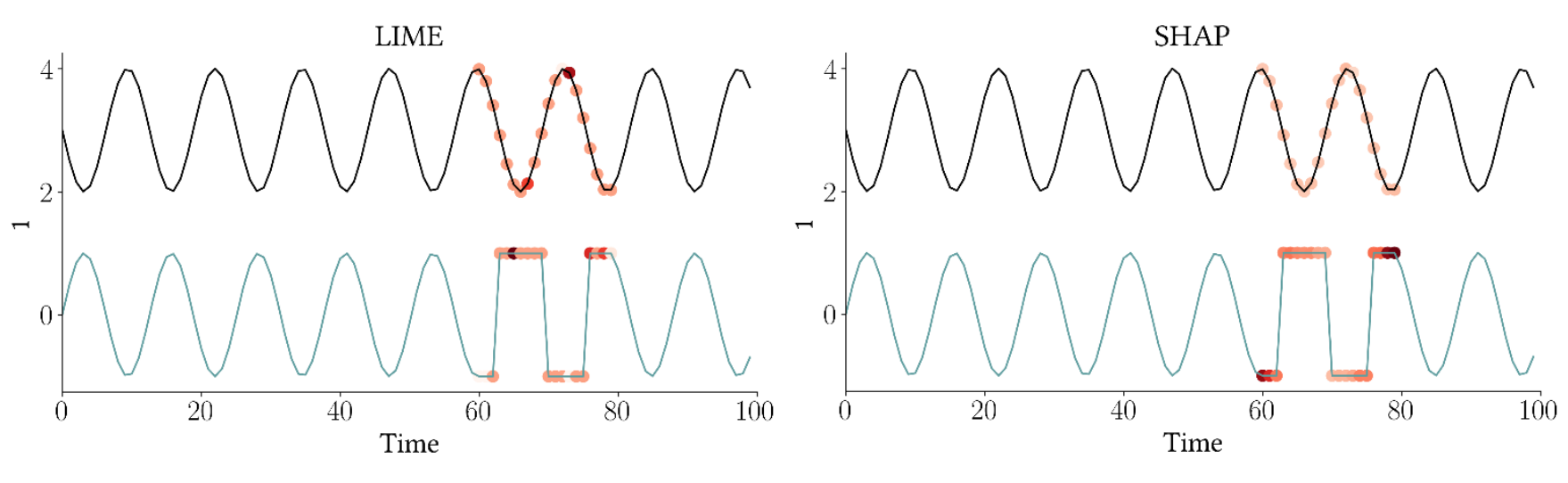}
	\caption{XEM with LIME and SHAP feature importance results from an MTS of the synthetic dataset belonging to the \textit{positive} class.}
	\label{fig:shap_synthetic_intepretability}
	\vspace{-1em}
\end{figure}

The classification results show that XEM with a time window size parameter set to 20\% is enough to correctly classify the 10 MTS of the test set (accuracy: 100\% - $n\_trees$: 10, $max\_depth$: 1). Moreover, the classification results for the \textit{positive} class MTS are based on the 20\% time window with a square signal on dimension 1. We observe that the maximum class probability for the MTS of \textit{positive} class is 100\% and this probability is reached for samples on the range [62,100] (maximum class probability on the range [0,61]: 92.6\%). This range is the expected range. As explained in section~\ref{Dataset_Transformation}, all the samples of the dataset obtained with a 20\% sliding window have a piece of the square signal for the timestamps in the range [62,100], which is the information sufficient to correctly classify the MTS in the \textit{positive} class. 
Furthermore, a time window size set below 20\% also leads to 100\% accuracy on the test set as a piece of the square signal (20\% of the MTS) is enough to correctly classify the MTS of the \textit{positive} class. For example, using the minimum window size (2\%), we observe that the maximum class probability obtained by XEM (accuracy: 100\% - $n\_trees$: 10, $max\_depth$: 1) for the MTS of \textit{positive} class is 100\% and this probability is reached for samples in the range [61,81] (maximum class probability on the range [0,60] and [82,100]: 97.8\%). This is also the expected discriminative range. Therefore, XEM can classify an MTS based on the minimal discriminative window; and by taking all the samples of the dataset with the maximum class probability, XEM can identify the full parts of the MTS which are characteristic of a class (e.g., the square signal on 20\% of the dimension 1 in Figure~\ref{fig:synthetic_intepretability}).\\
Then, we compare XEM explainability-by-design results with the ones from the post hoc model-agnostic explainability methods LIME and SHAP applied to XEM. Figure~\ref{fig:shap_synthetic_intepretability} shows the results from LIME and SHAP for a sample belonging to the \textit{positive} class, with the darker the red color the higher the importance to the predictions. 
First, we can see that, unlike XEM explainability-by-design (see Figure~\ref{fig:synthetic_intepretability}), LIME and SHAP do not homogeneously identify the discriminative square signal in Dimension 1 (interval [60,80]) as important to the prediction. SHAP identifies the timestamps at the beginning and at the end of the discriminative window as more important to the prediction than the other ones, therefore explaining to the end-user that the interval [65,75] is less discriminative to the prediction, which is not the expected result. A comparable observation can be made on LIME results. Second, LIME and SHAP provide some non-null importance values for the Dimension 2, which is not discriminative as the sine wave is common to both classes, therefore generating a misleading explanation for the end-user. Thus, this example, based on the same XEM model and a known ground truth with regard to the expected explanation, emphasizes that the explanations coming from the surrogate models of some post hoc model-agnostic explainability methods like LIME and SHAP are not perfectly faithful, and demonstrates the interest to have the combination of performance and explainability-by-design of XEM which provides the discriminative time window as explanation.

\begin{figure}[!htpb]
	\centering
	\includegraphics[width=.5\linewidth]{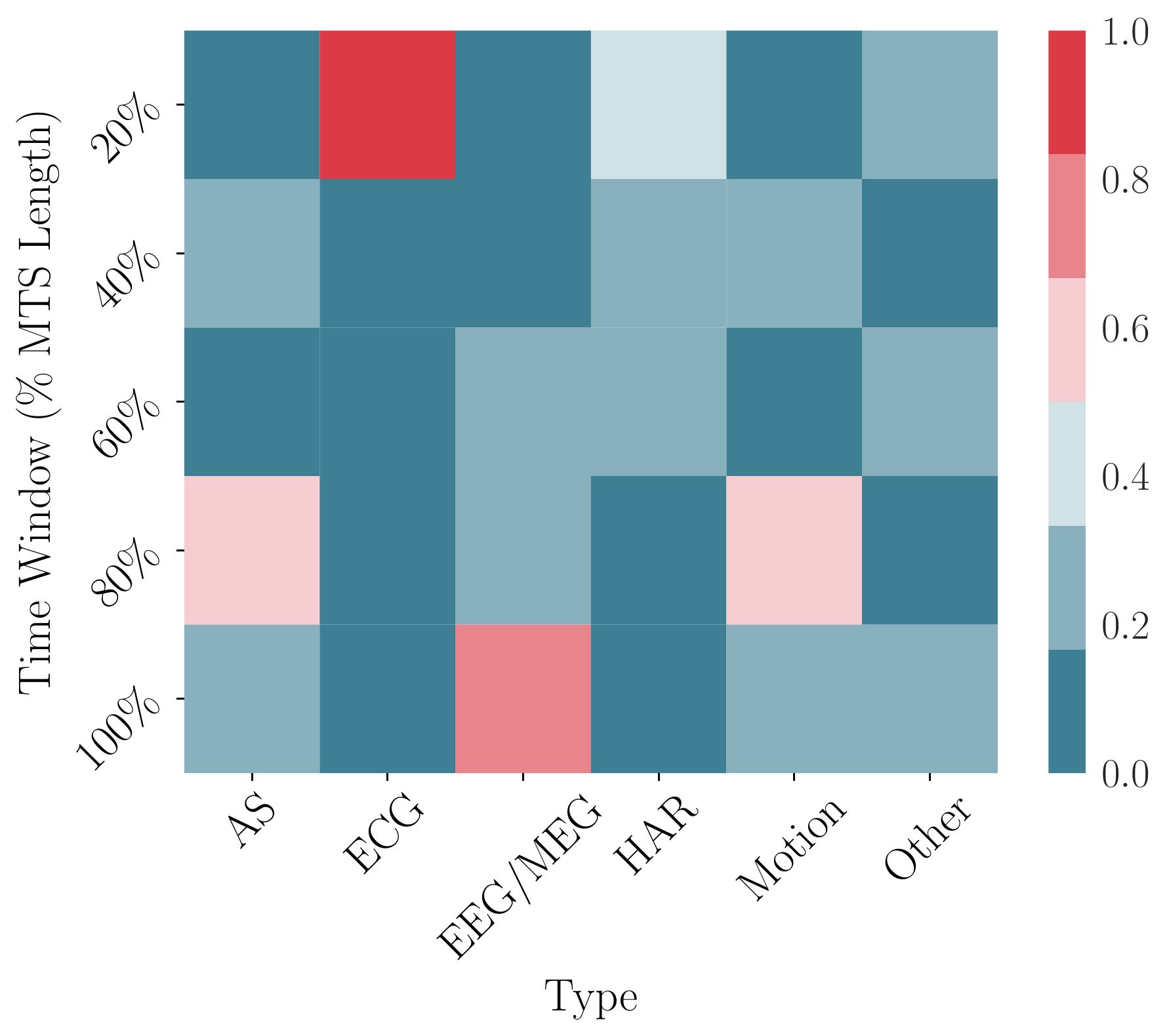}
	\caption{Heatmap of the proportion of the time window size percentages ($win\_size$) used by XEM per UEA dataset type.}
	\label{fig:heatmap_intepretability}
	\vspace{-1em}
\end{figure}

\textbf{Time Window Size Percentages on UEA} We then present the XEM explainability results on the UEA datasets. We begin with illustrating in Figure \ref{fig:heatmap_intepretability} the distribution of the time window size percentage used by XEM on the UEA archive per dataset type. We observe that XEM has a tendency to use particular time window size percentages per dataset type. Most of audio spectra, EEG/MEG and motion datasets have been classified on a time window size $> 60\%$ of the MTS lengths. Meanwhile, most ECG and human activity recognition datasets have been classified on a time window size $\leq 60\%$ of the MTS lengths. Therefore, we can induce that the information provided by the whole MTS is useful to discriminate between the different classes on the audio spectra, EEG/MEG and motion datasets. Concerning the ECG and human activity recognition datasets, we can infer that the discriminative information is located in a particular part of the MTS.

\begin{figure*}[!htpb]
	\centering
	\includegraphics[width=\linewidth]{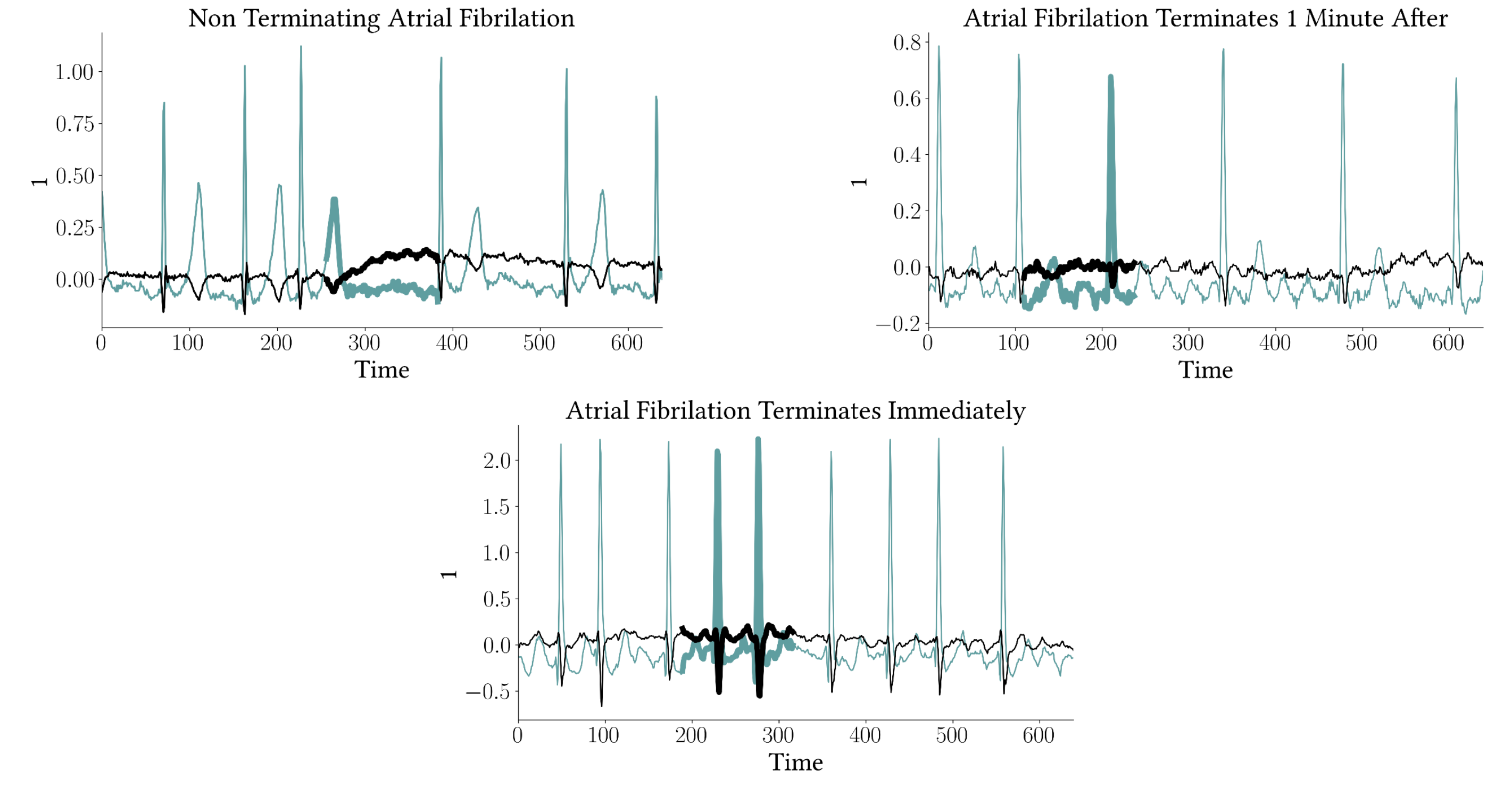}
	\caption{First MTS sample per class of Atrial Fibrilation test set with the XEM time window used for classification highlighted in bold, which serves as explanation for the end-user ($win\_size$: 20\%).}
	\label{fig:graph_intepretability}
	\vspace{-1em}
\end{figure*}

\textbf{Atrial Fibrilation Dataset} For example, XEM obtains its best performance on the two ECG datasets using a time window size of 20\%. Therefore, we assume that the information necessary for XEM to classify the MTS in ECG datasets are really condensed compared to the entire MTS available. We illustrate it in Figure~\ref{fig:graph_intepretability} by highlighting the 20\% time window of the first MTS sample per class in the Atrial Fibrilation test set to gain insights on XEM classification result. Atrial Fibrilation dataset is composed of two channels ECG on a 5 second period (128 samples per second). MTS are labeled in 3 classes: {\em non-terminating atrial fibrilation}, {\em atrial fibrilation terminates one minute after} and {\em atrial fibrilation terminates immediately}. XEM correctly predicts the 3 MTS based on the one second time window (20\%) highlighted in Figure~\ref{fig:graph_intepretability}. There is a unique window for each MTS with the highest class probability (class {\em non-terminating atrial fibrilation}: 94.6\%, {\em atrial fibrilation terminates one minute after}: 97.7\%, {\em atrial fibrilation terminates immediately}: 97.4\%). We can observe in the {\em non-terminating atrial fibrilation} MTS that the time window highlighted reveals an abnormal constant increase on channel 2 (black line) during one second whereas the other channel keeps the same motif as other windows. On the {\em atrial fibrilation terminates one minute after} MTS, we observe a smaller decrease in channel 2 than in other windows and a low peak in channel 1. These particular 20\% time windows inform the end-user about XEM classification outcome, thus providing important information to domain experts.

\begin{figure}[!htpb]
	\centering
	\includegraphics[width=.95\linewidth]{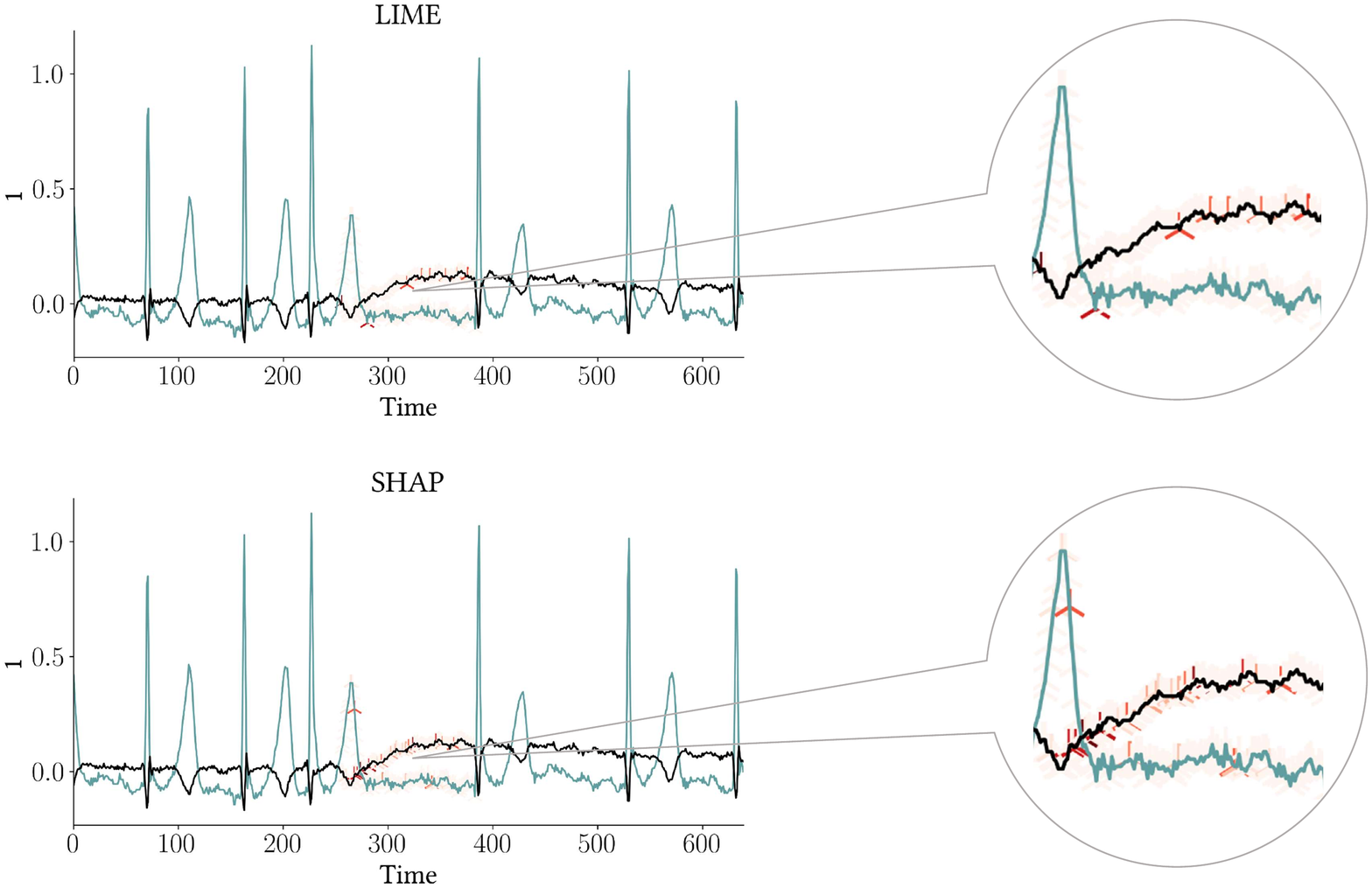}
	\caption{XEM with LIME and SHAP feature importance results from the first MTS of the Atrial Fibrilation test set belonging to the \textit{non-terminating atrial fibrilation} class.}
	\label{fig:shap_af_intepretability}
	\vspace{-1em}
\end{figure}

Then, we also compare XEM explainability-by-design results presented in Figure~\ref{fig:graph_intepretability} with the ones from the post hoc model-agnostic explainability methods LIME and SHAP applied to XEM. Figure~\ref{fig:shap_af_intepretability} shows the results from LIME and SHAP for a sample belonging to the \textit{non-terminating atrial fibrilation} class, with the darker the red color the higher the importance to the predictions. As observed on the synthetic dataset, the regions with high importance provided by LIME and SHAP are discontinued on channel 2, rendering it difficult for the end-user to interpret this explanation. Moreover, for both LIME and SHAP, only one or two points are identified as important on channel 1 without a clear interpretation associated to them, and can therefore be considered as noise. This example also supports the interest of XEM explainability-by-design which provides the discriminative time window as explanation.

\begin{figure*}[!htpb]
	\centering
	\includegraphics[width=.95\linewidth]{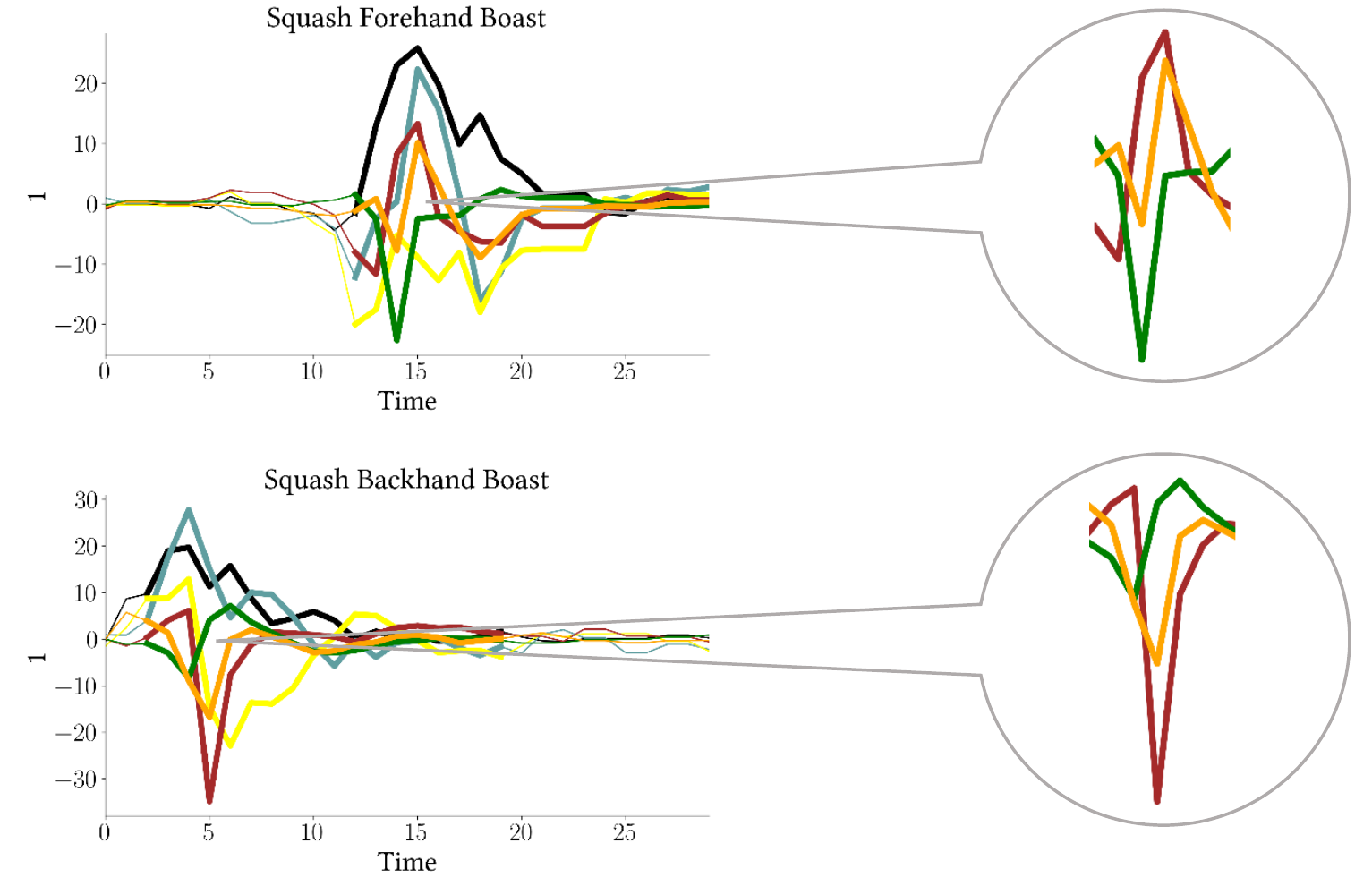}
	\caption{First MTS sample per class of Squash Racket Sports test set with the XEM time window used for classification highlighted in bold, which serves as explanation for the end-user ($win\_size$: 60\%).}
	\label{fig:graph_intepretability_racketsports}
\end{figure*}

\textbf{Racket Sports Dataset} The second category of datasets where XEM obtains its best results on a time window size $\leq 60\%$ of the MTS lengths is human activity recognition. As previously done with Atrial Fibrilation, we illustrate it in Figure~\ref{fig:graph_intepretability_racketsports} by highlighting the 60\% time window of the first MTS sample per class in the Racket Sports test set to gain insights on XEM classification result. Racket Sports dataset is composed of 6 dimensions, x/y/z coordinates for both the gyroscope and accelerometer of an android phone, on a 3 second period (10 samples per second). MTS are labeled in 4 classes: {\em badminton smash}, {\em badminton clear}, {\em squash forehand boast} and {\em squash backhand boast}. We illustrate the explainability of XEM on the two classes relative to the squash: {\em squash forehand boast} and {\em squash backhand boast}. XEM correctly predicts the 2 MTS based on the 1.8 seconds time window (60\%) highlighted in Figure~\ref{fig:graph_intepretability}. There is a unique window for each MTS with the highest class probability ({\em squash forehand boast}: 90.3\%, {\em squash backhand boast}: 86.7\%). We can observe that for these 2 MTS the window highlighted well correspond to the period of the full movement. Then, we can see a simultaneous steep peak on red and orange dimensions with a steep decrease on green dimension for {\em squash forehand boast}. Whereas, we can see a simultaneous steep decrease on red and orange dimensions without a particular variation on the green dimension for {\em squash backhand boast}. These particular 60\% time windows inform the end-user about XEM classification outcome, thus providing important information to domain experts.

\begin{figure}[!htpb]
	\centering
	\includegraphics[width=\linewidth]{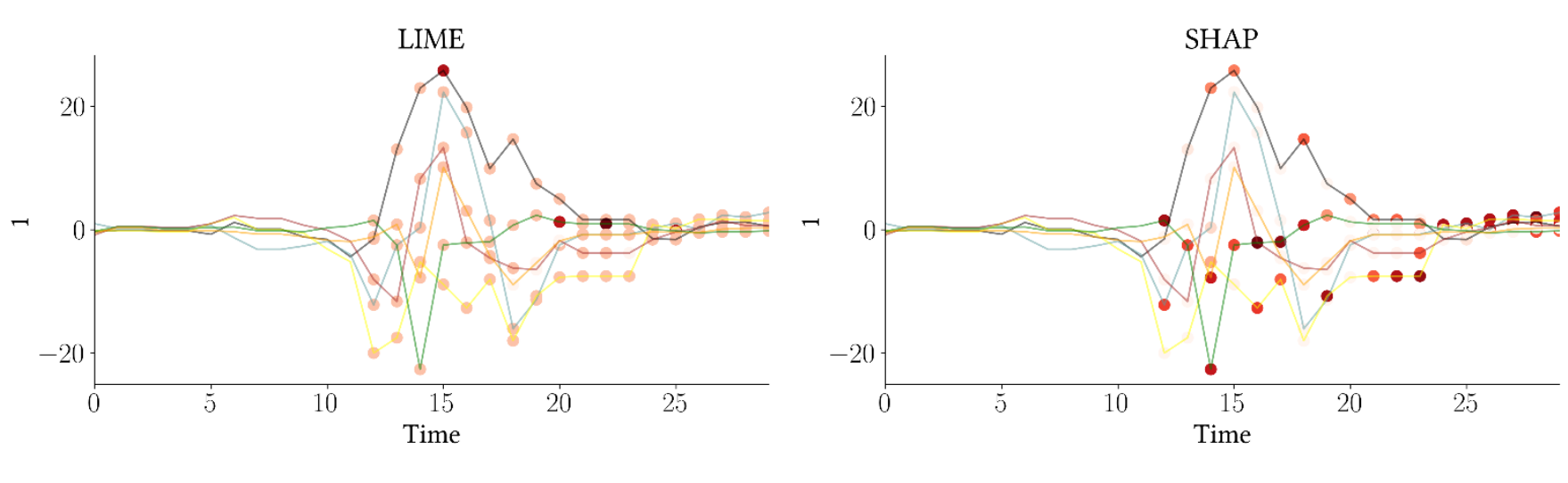}
	\caption{XEM with LIME and SHAP feature importance results from the first MTS of the Racket Sports test set belonging to the \textit{squash forehand boast} class.}
	\label{fig:shap_rs_intepretability}
	\vspace{-1em}
\end{figure}

Finally, we compare XEM explainability-by-design results presented in Figure~\ref{fig:graph_intepretability_racketsports} with the ones from the post hoc model-agnostic explainability methods LIME and SHAP applied to XEM. Figure~\ref{fig:shap_rs_intepretability} shows the results from LIME and SHAP for a sample belonging to the \textit{squash forehand boast} class, with the darker the red color the higher the importance to the predictions. As observed on the synthetic dataset, LIME and SHAP results only identify part of the discriminative features (e.g., do not identify steep peak on red and orange dimensions) and put some importance on non relevant parts of the time series (e.g., most of high LIME and SHAP importance values are after timestamp 18 - when the movement is finished). Such observations underline the imperfect faithfulness limitation of some post hoc model-agnostic explainability methods like LIME and SHAP, and the interest for XEM explainability-by-design. Nonetheless, XEM explainability-by-design faces some limitations coming from the use of a fixed-length time window, and these limitations are discussed in section~\ref{discussion}.

These two examples show how XEM outperforms other MTS classifiers (rank 1 on Atrial Fibrilation and Racket Sports) while offering faithful explainability-by-design on its predictions. 

\subsubsection{Effect of Missing Data}
\label{Missing_Data}
None of the state-of-the-art MTS classifiers handles missing data. Missing data are interpolated, which adds a parameter to the problem. Similar to extreme gradient boosting~\citep{Chen16}, XEM excludes missing values for the split and uses block propagation. 
Block propagation sends all samples with missing data to the node maximizing the accuracy score.

\begin{figure}[!htpb]
	\centering
	\includegraphics[width=\linewidth]{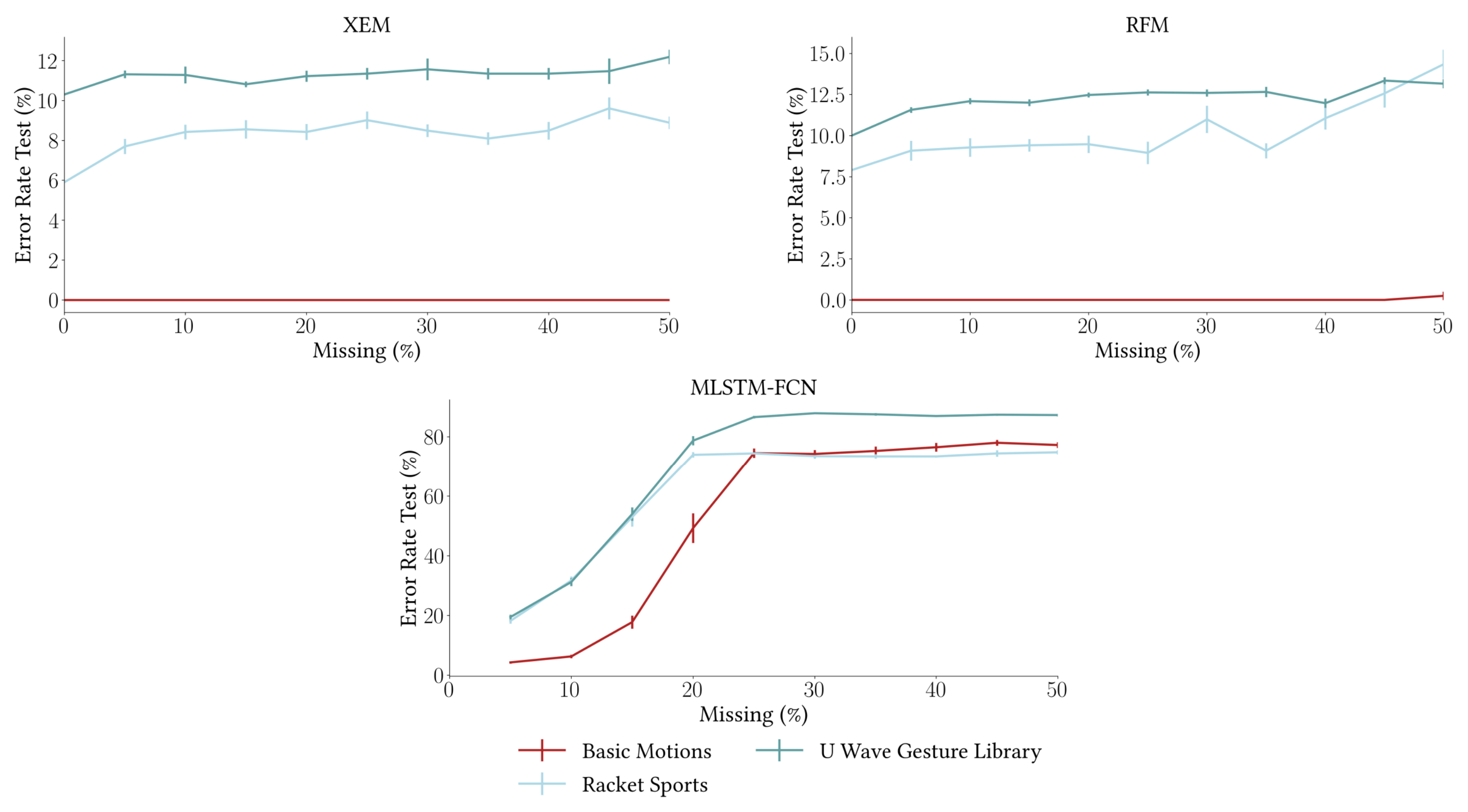}
	\caption{Evolution of XEM, RFM and MLSTM-FCN error rates with standard errors according to the proportion of missing values on three Human Activity Recognition datasets.}
	\label{fig:Missing}
\end{figure}

We present in this section an experiment to illustrate the performance of XEM in the case of missing data compared to the second and third ranked MTS classifiers (RFM and MLSTM-FCN - see Table~\ref{tab:UEA}) with an imputation method for missing values. We have selected three datasets from the most representing type of UEA datasets (human activity recognition, 30\% of the datasets); it is also a type on which XEM does not obtain the best performance comparing to the other classifiers (rank: 3.6). We choose the three datasets according to the performance of XEM to show the evolution of accuracies according to different starting points: Basic Motions (XEM accuracy: 100\%, no error), Racket Sports (94.1\%, ]0,10] percent of error) and U Wave Gesture Library (89.7\%,  ]10,100] percent of error). Then, we randomly removed an increasing proportion of the values for each time series ($\{5\%, 10\%, ... , 50\%\}$) of the datasets before transformation (see section~\ref{Dataset_Transformation}). For RFM and MLSTM-FCN, missing values are filled with zeros. Classifiers are trained following the methodology described in section~\ref{Evaluation} and the error rates on test sets over 10 replications are presented in Figure~\ref{fig:Missing}.

First, we observe that missing data does not have an effect on XEM performance (100\% accuracy) on the dataset Basic Motions. On the other two datasets, the error rates of XEM increase progressively with the proportion of missing data. The error rate induced by missing data never exceeds 5\% on these 2 datasets when half the data is missing (accuracy difference from 0\% to 50\% missing data: Racket Sports +3.7\% and U Wave Gesture Library +1.9\%). Finally, XEM performance is stable: the error rates remain roughly the same across the 10 replications on all proportions of missing values (mean of standard error across Racket Sports/U Wave Gesture Library: 0.34\%).

Then, we can see that missing data has a stronger effect on RFM classification performance than XEM on the three datasets (error difference from 0\% to 50\% missing data: Basic Motions +0.25\% versus 0\%, Racket Sports +6.4\% versus +3.7\% and U Wave Gesture Library +3.2\% versus +1.9\%). Nonetheless, the effect of missing data on XEM and RFM performance remains below 10\%. This observation is not applicable to MLSTM-FCN which is highly impacted by the missing data. MLSTM-FCN performance drops sharply on all datasets and it is not able to learn anymore from the data when the proportion of missing values exceed 25\% (same performance as a random classifier - accuracy of one over the number of classes). Considering that MLSTM-FCN and RFM have the same imputation method, we can assume that using the window on which RFM is the most confident for prediction confers a higher robustness to missing values.

Therefore, this experiment highlights the interest of classifying based on the subsequence on which XEM is the most confident and the advantage conferred by its natural way to handle missing values compared to its competitors.

\subsubsection{Effect of Gaussian Noise}
\label{Noise}
In this section, we evaluate the robustness of XEM to Gaussian noise compared to the second and third ranked MTS classifiers. Therefore, we compare the performance of XEM to RFM and MLSTM-FCN, with RFM proven to be robust to noise based on bagging~\citep{Breiman96}.

\begin{figure}[!htpb]
	\centering
	\includegraphics[width=.6\linewidth]{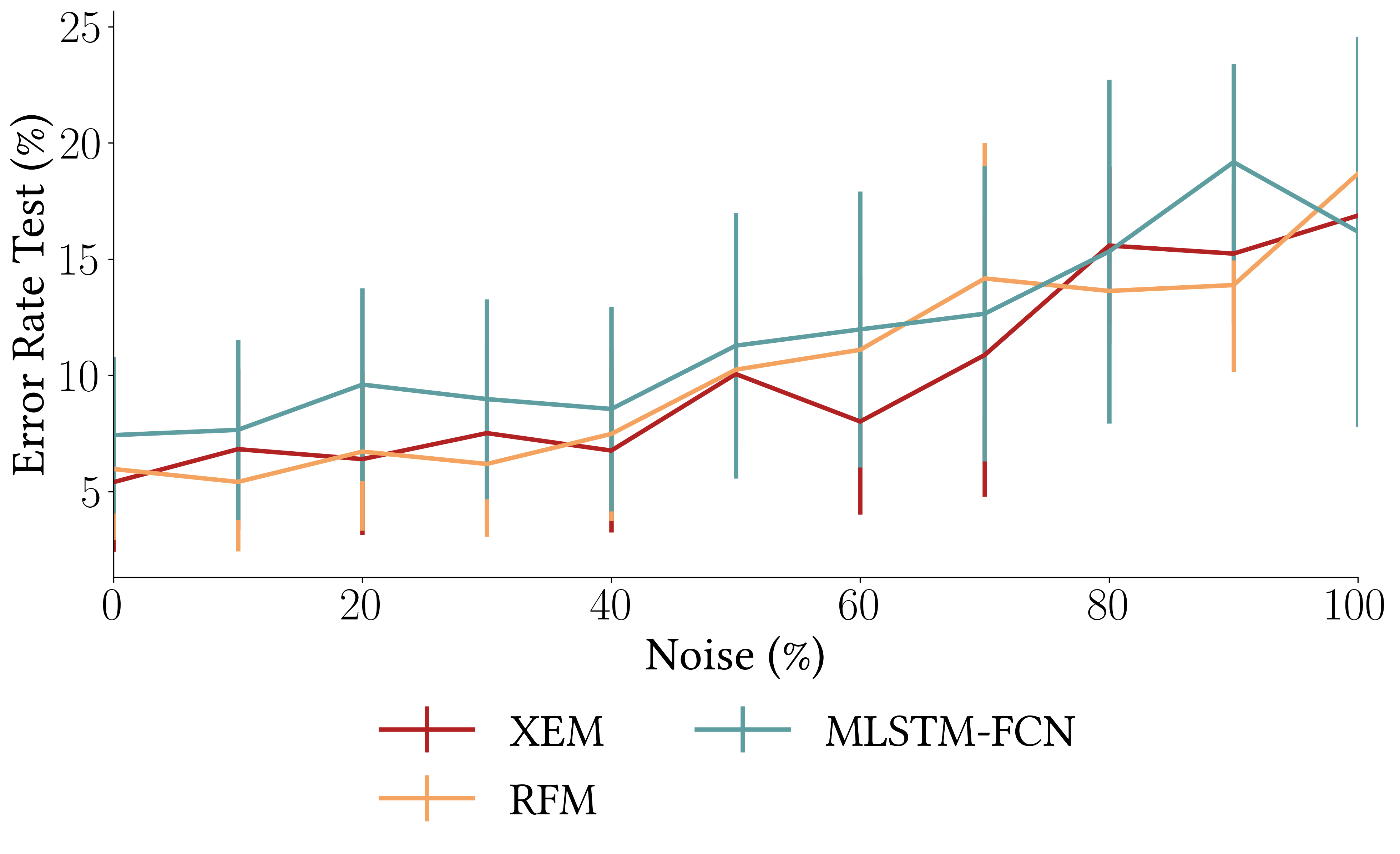}
	\caption{Evolution of the top three MTS classifiers average error rates with standard errors on three Human Activity Recognition datasets (Basic Motions, Racket Sports, U Wave Gesture Library) according to the level of noise.}
	\label{fig:Noise}
	\vspace{-1em}
\end{figure}

Following the same logic as the section on missing values, we performed an experiment on the same three datasets. These three datasets are from the most representing type of UEA datasets (human activity recognition, 30\% of the datasets) and from different XEM accuracy categories: Basic Motions (XEM accuracy: 100\%, no error), Racket Sports (94.1\%, ]0,10] percent of error) and U Wave Gesture Library (89.7\%, ]10,100] percent of error). Then, after z-normalization of these datasets on each dimension (standard deviation of 1), we added an increasing Gaussian noise with a standard deviation of 0 to 1 to each dimension, which is equivalent to noise levels of 0\% to 100\%. The average error rates with standard errors on these three datasets are presented in Figure~\ref{fig:Noise}.

We observe that XEM fully exploits its bagging component and is as robust to noise as RFM. XEM shows lower error rates than RFM on 60\% of the noise levels, without having a greater variability across the datasets (average standard error: XEM 3.7\% versus RFM 3.5\%). Moreover, XEM is more robust to noise than MLSTM-FCN. XEM exhibits lower error rates than MLTSM-FCN on 80\% of the noise levels with a lower variability across the datasets (average standard error: XEM 3.7\% versus MLSTM-FCN 5.3\%).

\subsubsection{Performance-Explainability Framework}
\label{framework}
As previously presented, XEM is the first MTS classifier reconciling performance and faithful explainability.
In this section, we position XEM in the performance-explainability framework~\citep{Fauvel20Framework} in comparison with the state-of-the-art MTS classifiers (DTW$_{I}$/DTW$_{D}$, MLTSM-FCN and WEASEL+MUSE), and identify ways to further enhance XEM explainability.

\begin{figure*}[!htpb]
	\centering
	\includegraphics[width=\linewidth]{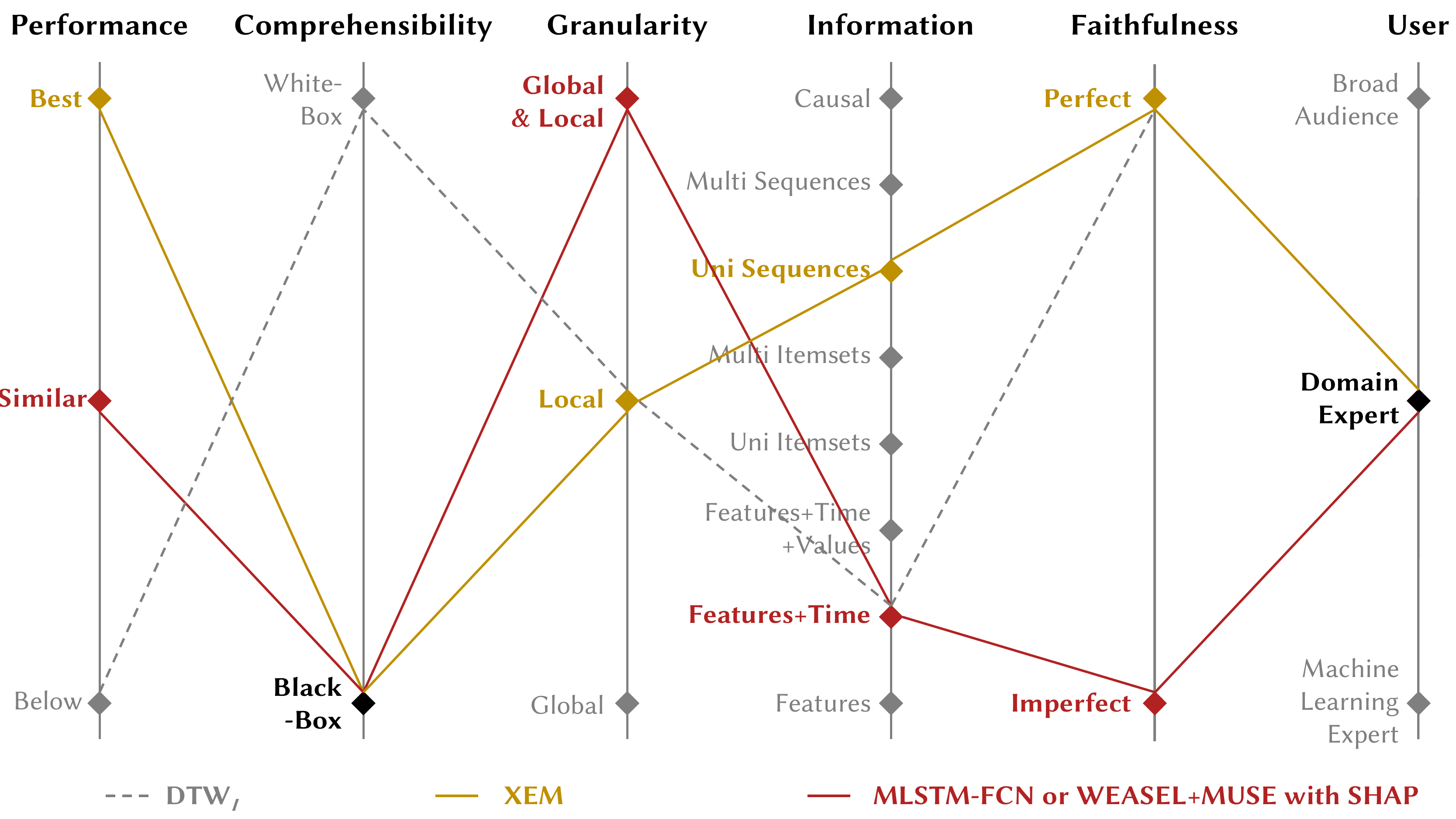}
	\caption{Parallel coordinates plot of XEM and the state-of-the-art MTS classifiers. Performance evaluation method: predefined train/test splits and an arithmetic mean of the accuracies on the 30 public UEA datasets~\citep{Bagnall18}. As presented in section~\ref{RW_MTSClassification}, the models evaluated in the benchmark are: DTW$_{D}$, DTW$_{I}$, FCN, gRSF, LPS, MLSTM-FCN, mv-ARF, ResNet, SMTS, TapNet, UFS, WEASEL+MUSE and XEM.}
	\label{fig:Framework}
	\vspace{-1em}
\end{figure*}

The performance-explainability framework details a set of 6 characteristics (performance, model comprehensibility, granularity of the explanations, information type, faithfulness and user category) to assess and benchmark machine learning methods.
The results of the framework are represented in a parallel coordinates plot in Figure~\ref{fig:Framework}.

Firstly, DTW$_{I}$ classifies MTS samples based on the label of their nearest sample. The similarity is calculated as the cumulative distances of all dimensions independently measured under DTW. For an individual MTS, the explanations supporting the prediction are the ranking of features and timestamps in decreasing order of their DTW distance with the nearest MTS. Based on the results presented in section~\ref{XEM_Performance}, DTW$_{I}$ underperforms the current state-of-the-art MTS classifiers (Performance: \textit{Below}) as it has a statistically significant lower performance than MLSTM-FCN. In addition, the model DTW$_{I}$ conveys limited information (Information: \textit{Features+Time}) that needs to be analyzed by a domain expert to ensure that they are meaningful for the application (User: \textit{Domain Expert}). However, DTW$_{I}$ model is comprehensible (Comprehensibility: \textit{White-Box}) and provides faithful explanations (Faithfulness: \textit{Perfect}) for each MTS (Granularity: \textit{Local}).

Then, MLTSM-FCN and WEASEL+MUSE can be analyzed together. 
First, based on the results presented in section~\ref{XEM_Performance}, MLSTM-FCN exhibits the third best performance followed by WEASEL+MUSE without showing a statistically significant performance difference with XEM (Performance: \textit{Similar}). 
Second, they are both ``black-box'' classifiers without providing explainability-by-design or, as far we have seen, having a post hoc model-specific explainability method. Therefore, their explainability characteristics depend on the choice of the post hoc model-agnostic explainability method. 
Using the popular state-of-the-art post hoc model-agnostic explainability method SHAP, it allows WEASEL+MUSE and MLS- TM-FCN to outperform DTW$_{I}$ while reaching explanations with a comparable level of information (Information: \textit{Features+Time}, DTW$_{I}$: \textit{Features+Time}), in the meantime remaining accessible to a domain expert (User: \textit{Domain Expert}, DTW$_{I}$: \textit{Domain Expert}). However, as opposed to DTW$_{I}$, SHAP as a surrogate model does not provide perfectly faithful explanations (Faithfulness: \textit{Imperfect}, DTW$_{I}$: \textit{Perfect}).

Finally, XEM exhibits the best performance (Performance: \textit{Best}) while providing faithful (Faithfulness: \textit{Perfect}, MLSTM-FCN/WEASEL+MUSE: \textit{Imperfect}, DTW$_{I}$: \textit{Perfect}) and more informative explanations as it provides the time window used to classify the whole MTS (Information: \textit{Uni Sequences}, MLSTM-FCN/ WEASEL+MUSE: \textit{Features+Time}, DTW$_{I}$: \textit{Features+Time}).
However, the explanations supporting XEM predictions are only available per MTS (Granularity: \textit{Local}) and the level of information could be further enhanced. It would be interesting to analyze the time windows characteristic of each class in the training set in order to determine if they contain some common multidimensional sequences (Information: \textit{Multi Sequences}, Granularity: \textit{Both Global \& Local}). Such patterns could also broaden the audience as they would synthesize the important information in the discriminative time windows.

\subsection{Discussion}
\label{discussion}
We have presented our new eXplainable Ensemble method for MTS classification (XEM), which relies on the new hybrid ensemble method LCE. We have shown that LCE outperforms the state-of-the-art classifiers on the UCI datasets and that XEM outperforms the state-of-the-art MTS classifiers on the UEA datasets. In addition, XEM provides faithful explainability-by-design and manifests robust performance when faced with challenges arising from continuous data collection (different MTS length, missing data and noise). However, our new method XEM has some limitations due to the use of a fixed-length time window to classify an MTS.

\begin{figure*}[!htpb]
	\centering
	\includegraphics[width=\linewidth]{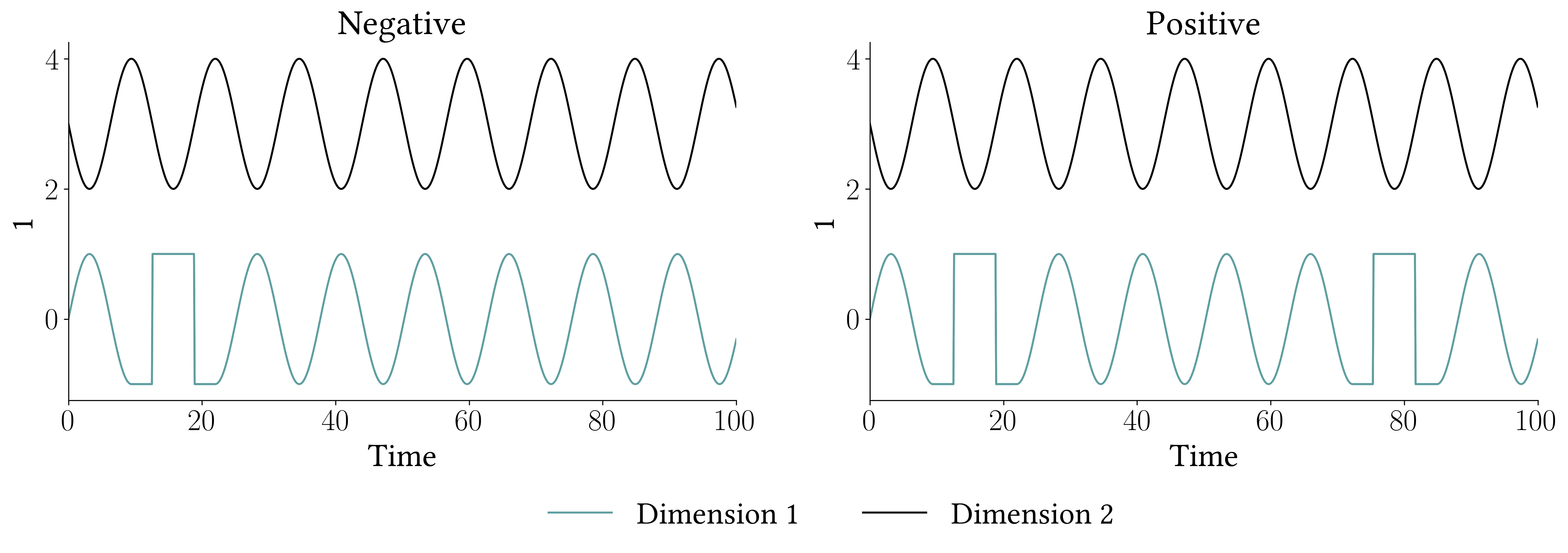}
	\caption{Two MTS samples of the synthetic dataset with the \textit{positive} class having two square signals.}
	\label{fig:Discussion_Synthetic}
	\vspace{-1em}
\end{figure*}

\begin{figure*}[!htpb]
	\centering
	\includegraphics[width=\linewidth]{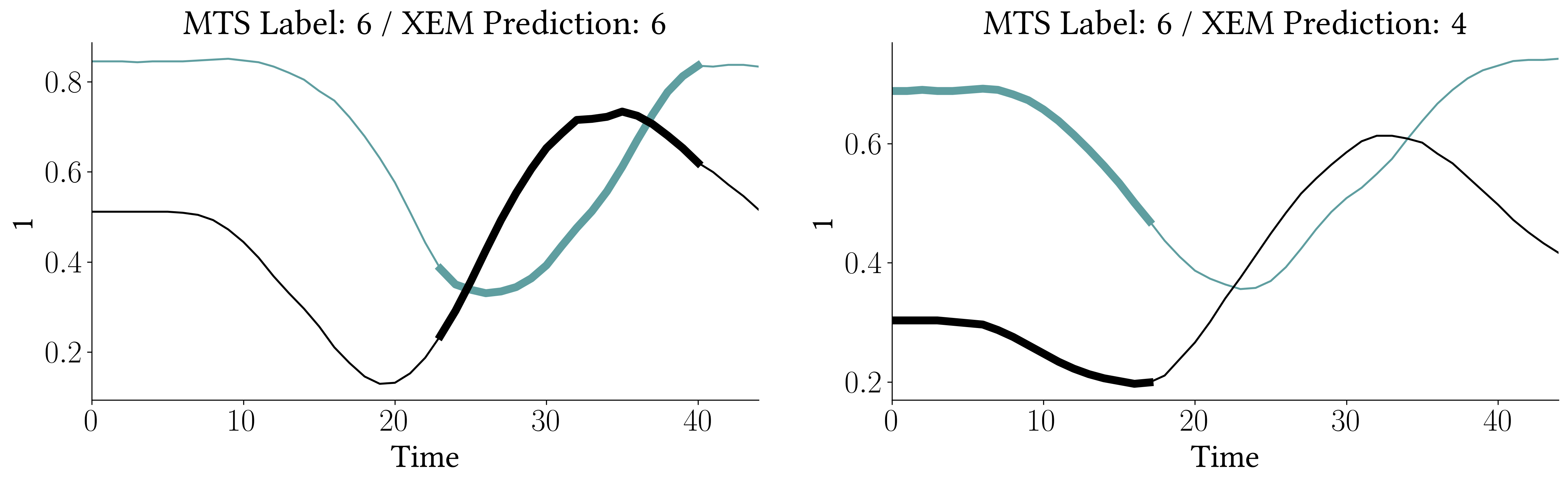}
	\caption{Two MTS samples of Libras test set belonging to the same class with XEM predictions and the time windows used for classification highlighted in bold, which serves as explanation for the end-user ($win\_size$: 40\%).}
	\label{fig:Discussion}
	\vspace{-1em}
\end{figure*}

Firstly, some limitations arise from the consideration of only one window. Depending on the dataset, XEM can face (\textit{i}) a drop in the precision of the explanation or (\textit{ii}) fail to identify the discriminant window. Specifically, (\textit{i}) XEM can face a drop in the precision of the explanation in case of discriminative features located on non-consecutive time windows. The precision can be defined as the fraction of explanations that is relevant to the prediction. XEM uses one window to identify the discriminative part of an MTS. Nevertheless, some MTS can be solely distinguished based on the combination of several non-consecutive time windows. In this case, in order to include all discriminative information to correctly perform the classification, XEM selects a time window covering all necessary non-consecutive time windows, therefore altering the precision of the explanation provided to the end-user by including some unnecessary information. We illustrate this scenario based on the experiment performed on the synthetic dataset in section~\ref{Res_Explainability}. Here, the difference between the 10 MTS belonging to the \textit{negative} class and the one belonging to the \textit{positive} class stems from the presence of one (t1=9) or two square signals (t1=9 and t2=72, see Figure~\ref{fig:Discussion_Synthetic}). Each square signal is of size 12, therefore a window size below 63\% (72-9) does not allow the discrimination between the two MTS classes; a window covering both square signals is necessary to perform this task. We observe that XEM correctly identifies the discriminative time window by obtaining a 100\% accuracy using an 80\% time window (accuracy of 50\% with a time window in $\{20\%, 40\%, 60\%\}$). Nonetheless, the explanation communicated to the end-user (80\% time window [5,85]) to support the prediction contains 56 timestamps which are not relevant - sine wave, inducing a drop in the precision of the explanation (precision: 30\% = 2*12/80 versus 100\% in the case of consecutive discriminative parts). Therefore, to circumvent this limitation, it would be interesting to develop a method that would synthetize under the form of patterns the time windows characteristic of each class (as suggested in the previous section as well), and so provide to the end-user solely the discriminative parts of an MTS as explanation.
In addition, \textit{(ii)} XEM can fail to identify the discriminative time window in case of a window with high proximity to another class. XEM predicts the class of an MTS based on the window on which it is the most confident, without considering the predictions on the other windows. Some datasets can contain MTS with different windows close to the characteristics of different classes. Therefore, XEM can have high class probabilities on multiple windows; and when the window on which XEM is the most confident is characteristic of another class than the expected one, XEM incorrectly classifies the MTS. To illustrate it, we present in Figure 14 two MTS of the UEA Libras test set. XEM performed poorly on this dataset and obtained the rank 10/11 (see section~\ref{XEM_Performance}). The Libras dataset contains 15 classes of 24 instances each, where each class references a hand movement type in the Brazilian sign language Libras. The hand movement is represented as a bi-dimensional curve performed by the hand in a period of time. We can observe in Figure~\ref{fig:Discussion} that the two MTS belonging to the same class have comparable evolution across time but XEM classifies them into two different classes. The first MTS is correctly classified based on the time window [23,40] with a class probability equals to 93.5\%. We can assume that the evolution on this window is characteristic of the class 6 (\textit{circle movement}). The second MTS also contains a comparable window on the range [23,40] but is incorrectly classified based on another window (range [0,17]) with a class probability of 94.5\%. Therefore, XEM is the most confident on a window characteristic of another class (class 4: \textit{anti-clockwise arc}). XEM did not consider the predictions on the other windows to take its decision. More particularly, XEM did not consider the expected window [23,40], where it also gets a high-class probability of 86.3\%. So, it would be interesting to improve our hybrid ensemble method for MTS classification by considering in the final decision the predictions on the different windows of an MTS.

\begin{figure*}[!htpb]
	\centering
	\includegraphics[width=\linewidth]{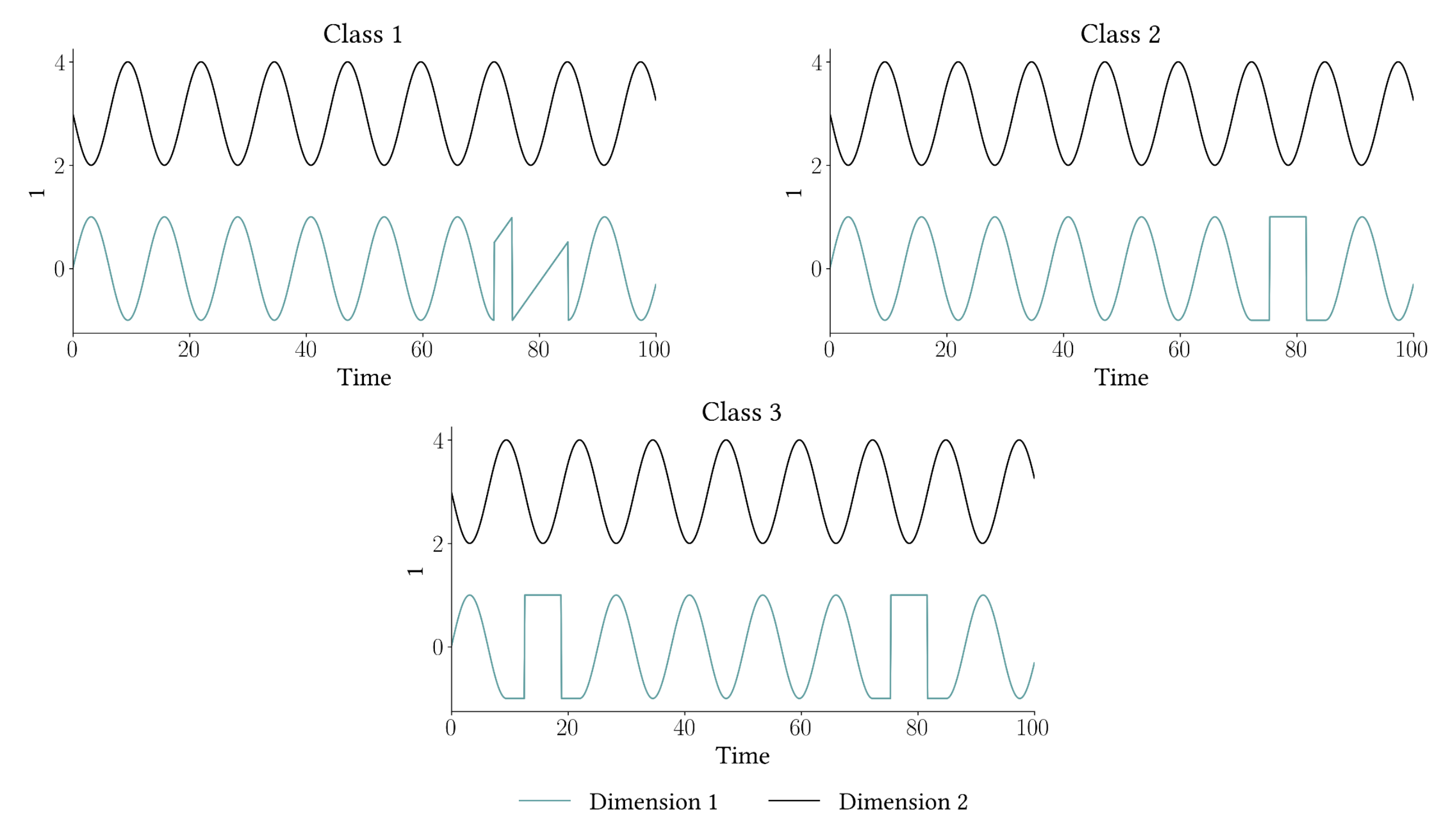}
	\caption{The three MTS types of the synthetic dataset.}
	\label{fig:Discussion_Synthetic2}
	\vspace{-1em}
\end{figure*}

Secondly, the choice of a time window with a fixed length can be another limitation. We assume in XEM that a unique window size is suitable to discriminate the different classes. Nonetheless, we can imagine that different classes can be characterized by signals of different lengths. This assumption leads XEM to select the window size associated with the class having the largest discriminative features, and affects the precision of explanation in case of other classes with smaller discriminative parts. For example, Figure~\ref{fig:Discussion_Synthetic2} shows an augmented version of the previous synthetic dataset (see Figure~\ref{fig:Discussion_Synthetic}) with a third class having a triangle wave in [72,84]. On this dataset and as seen in the previous example, XEM selects a window size of 80\% to correctly classify the different MTS, the window size covering both square signals of class 3. However, a window size of 20\% is sufficient to discriminate MTS from class 1 (triangle wave). Thus, given a window size of 80\% for this dataset, the explanation given to the end-user for an MTS sample belonging to the class 1 would contain information which is not discriminative (sine waves). Plus, adding some information/noise by taking a larger window than necessary for some of the classes can generate misclassifications for certain datasets. Therefore, it would also be valuable to improve XEM by integrating the possibility of multiple window sizes.

\section{Conclusion}
We have presented our new eXplainable-by-design Ensemble method for MTS classification (XEM), which relies on the new hybrid ensemble method LCE.
LCE exhibits a better average rank than the state-of-the-art classifiers on the public UCI datasets and XEM shows a better average rank than the state-of-the-art MTS classifiers on the public UEA datasets. As tree-based ensemble methods, LCE and XEM can scale well on larger datasets than the ones tested. In addition, XEM addresses the challenges MTS classification usually faces. First, it provides faithful explainability-by-design through the identification of the time window used to classify the whole MTS. Then, XEM is robust when faced with challenges arising from continuous data collection (different MTS length, missing data and noise). 

With regard to future work, we would like to adapt XEM approach to the regression task and evaluate it against the state-of-the-art regression methods. 
To further improve the explainability of XEM, we also plan to work on a method that would analyze the time windows characteristic of each class in the training set to determine if they contain some common multidimensional sequential patterns. Such patterns would enhance the level of information, the granularity of explanations (both global \& local) and could also broaden the audience as they would synthesize the important information in the discriminative time windows.

\section*{Acknowledgments}
This work was supported by the French National Research Agency under the Investments for the Future Program (ANR-16-CONV-0004) and the Inria Project Lab Hybrid Approaches for Interpretable AI (HyAIAI).

\bibliographystyle{plainnat}
\bibliography{References.bib}

\begin{thebibliography}{55}
\providecommand{\natexlab}[1]{#1}
\providecommand{\url}[1]{\texttt{#1}}
\expandafter\ifx\csname urlstyle\endcsname\relax
  \providecommand{\doi}[1]{doi: #1}\else
  \providecommand{\doi}{doi: \begingroup \urlstyle{rm}\Url}\fi

\bibitem[Bagnall et~al.(2018)Bagnall, Lines, and Keogh]{Bagnall18}
A.~Bagnall, J.~Lines, and E.~Keogh.
\newblock The {UEA UCR} {T}ime {S}eries {C}lassification {A}rchive.
\newblock 2018.

\bibitem[Baydogan and Runger(2014)]{Baydogan14}
M.~Baydogan and G.~Runger.
\newblock Learning a {S}ymbolic {R}epresentation for {M}ultivariate {T}ime
  {S}eries {C}lassification.
\newblock \emph{Data Mining and Knowledge Discovery}, 29\penalty0 (2):\penalty0
  400--422, 2014.

\bibitem[Baydogan and Runger(2016)]{Baydogan16}
M.~Baydogan and G.~Runger.
\newblock Time {S}eries {R}epresentation and {S}imilarity {B}ased on {L}ocal
  {A}utopatterns.
\newblock \emph{Data Mining and Knowledge Discovery}, 30\penalty0 (2):\penalty0
  476--509, 2016.

\bibitem[Bergstra et~al.(2011)Bergstra, Bardenet, Bengio, and
  K\'{e}gl]{Bergstra11}
J.~Bergstra, R.~Bardenet, Y.~Bengio, and B.~K\'{e}gl.
\newblock Algorithms for {H}yper-{P}arameter {O}ptimization.
\newblock In \emph{Proceedings of the 25th International Conference on Neural
  Information Processing Systems}, 2011.

\bibitem[Breiman(1996)]{Breiman96}
L.~Breiman.
\newblock Bagging {P}redictors.
\newblock \emph{Machine Learning}, pages 123--140, 1996.

\bibitem[Breiman(2001)]{Breiman01}
L.~Breiman.
\newblock Random {F}orests.
\newblock \emph{Machine Learning}, page 5–32, 2001.

\bibitem[Breiman et~al.(1984)Breiman, Friedman, Stone, and Olshen]{Breiman84}
L.~Breiman, J.~Friedman, C.~Stone, and R.~Olshen.
\newblock \emph{Classification and {R}egression {T}rees}.
\newblock The Wadsworth and Brooks-Cole statistics-probability series. Taylor
  \& Francis, 1984.

\bibitem[Chen and Guestrin(2016)]{Chen16}
T.~Chen and C.~Guestrin.
\newblock {XGB}oost: {A} {S}calable {T}ree {B}oosting {S}ystem.
\newblock In \emph{Proceedings of the 22nd ACM SIGKDD International Conference
  on Knowledge Discovery and Data Mining}, 2016.

\bibitem[Cussins~Newman(2019)]{Cussins19}
J.~Cussins~Newman.
\newblock Toward {AI} {S}ecurity: {G}lobal {A}spirations for a {M}ore
  {R}esilient {F}uture.
\newblock In \emph{Center for Long-Term Cybersecurity}, 2019.

\bibitem[Dem\v{s}ar(2006)]{Demsar06}
J.~Dem\v{s}ar.
\newblock Statistical {C}omparisons of {C}lassifiers over {M}ultiple {D}ata
  {S}ets.
\newblock \emph{Journal of Machine Learning Research}, 7:\penalty0 1--30, 2006.

\bibitem[Dietterich(2000)]{Dietterich00}
T.~Dietterich.
\newblock Ensemble {M}ethods in {M}achine {L}earning.
\newblock \emph{Multiple Classifier Systems}, pages 1--15, 2000.

\bibitem[Du et~al.(2020)Du, Liu, and Hu]{Du20}
M.~Du, N.~Liu, and X.~Hu.
\newblock Techniques for {I}nterpretable {M}achine {L}earning.
\newblock \emph{Communications of the ACM}, 2020.

\bibitem[Dua and Graff(2017)]{Dua17}
D.~Dua and C.~Graff.
\newblock {UCI} {M}achine {L}earning {R}epository, 2017.

\bibitem[Ebrahimpour et~al.(2012)Ebrahimpour, Sadeghnejad, Arani, and
  Mohammadi]{Ebrahimpour12}
R.~Ebrahimpour, N.~Sadeghnejad, S.~Arani, and N.~Mohammadi.
\newblock Boost-{W}ise {P}re-{L}oaded {M}ixture of {E}xperts for
  {C}lassification {T}asks.
\newblock \emph{Neural Computing and Applications}, 22\penalty0 (1):\penalty0
  365--377, 2012.

\bibitem[Esteva et~al.(2019)Esteva, Robicquet, Ramsundar, Kuleshov, DePristo,
  Chou, Cui, Corrado, Thrun, and Dean]{Esteva19}
A.~Esteva, A.~Robicquet, B.~Ramsundar, V.~Kuleshov, M.~DePristo, K.~Chou,
  C.~Cui, G.~Corrado, S.~Thrun, and J.~Dean.
\newblock A {G}uide to {D}eep {L}earning in {H}ealthcare.
\newblock \emph{Nature Medicine}, 25:\penalty0 24--29, 2019.

\bibitem[Fauvel et~al.(2019)Fauvel, Masson, Fromont, Faverdin, and
  Termier]{Fauvel19}
K.~Fauvel, V.~Masson, {\'E}.~Fromont, P.~Faverdin, and A.~Termier.
\newblock Towards {S}ustainable {D}airy {M}anagement - {A} {M}achine {L}earning
  {E}nhanced {M}ethod for {E}strus {D}etection.
\newblock In \emph{Proceedings of the 25th ACM SIGKDD International Conference
  on Knowledge Discovery and Data Mining}, 2019.

\bibitem[Fauvel et~al.(2020{\natexlab{a}})Fauvel, Balouek-Thomert, Melgar,
  Silva, Simonet, Antoniu, Costan, Masson, Parashar, Rodero, and
  Termier]{Fauvel20}
K.~Fauvel, D.~Balouek-Thomert, D.~Melgar, P.~Silva, A.~Simonet, G.~Antoniu,
  A.~Costan, V.~Masson, M.~Parashar, I.~Rodero, and A.~Termier.
\newblock A {D}istributed {M}ulti-{S}ensor {M}achine {L}earning {A}pproach to
  {E}arthquake {E}arly {W}arning.
\newblock In \emph{Proceedings of the 34th AAAI Conference on Artificial
  Intelligence}, 2020{\natexlab{a}}.

\bibitem[Fauvel et~al.(2020{\natexlab{b}})Fauvel, Masson, and
  Fromont]{Fauvel20Framework}
K.~Fauvel, V.~Masson, and {\'E}.~Fromont.
\newblock A {P}erformance-{E}xplainability {F}ramework to {B}enchmark {M}achine
  {L}earning {M}ethods: {A}pplication to {M}ultivariate {T}ime {S}eries
  {C}lassifiers.
\newblock In \emph{Proceedings of the IJCAI-PRICAI Workshop on Explainable
  Artificial Intelligence}, 2020{\natexlab{b}}.

\bibitem[Freund and Schapire(1996)]{Freund96}
Y.~Freund and R.~Schapire.
\newblock Experiments with a {N}ew {B}oosting {A}lgorithm.
\newblock In \emph{Proceedings of the 13th International Conference on Machine
  Learning}, 1996.

\bibitem[Gama and Brazdil(2000)]{Gama00}
J.~Gama and P.~Brazdil.
\newblock Cascade {G}eneralization.
\newblock \emph{Machine Learning}, 41\penalty0 (3):\penalty0 315--343, 2000.

\bibitem[Guidotti et~al.(2019)Guidotti, Monreale, Giannotti, Pedreschi,
  Ruggieri, and Turini]{Guidotti19}
R.~Guidotti, A.~Monreale, F.~Giannotti, D.~Pedreschi, S.~Ruggieri, and
  F.~Turini.
\newblock Factual and {C}ounterfactual {E}xplanations for {B}lack {B}ox
  {D}ecision {M}aking.
\newblock \emph{IEEE Intelligent Systems}, 34\penalty0 (6):\penalty0 14--23,
  2019.

\bibitem[He et~al.(2016)He, Zhang, Ren, and Sun]{He16}
K.~He, X.~Zhang, S.~Ren, and J.~Sun.
\newblock Deep {R}esidual {L}earning for {I}mage {R}ecognition.
\newblock In \emph{Proceedings of the 2016 IEEE Conference on Computer Vision
  and Pattern Recognition}, 2016.

\bibitem[Jacobs et~al.(1991)Jacobs, Jordan, Nowlan, and Hinton]{Jacobs91}
R.~Jacobs, M.~Jordan, S.~Nowlan, and G.~Hinton.
\newblock Adaptive {M}ixtures of {L}ocal {E}xperts.
\newblock \emph{Neural Computation}, 3\penalty0 (1):\penalty0 79–87, 1991.

\bibitem[Jiang et~al.(2019)Jiang, Song, Huang, Song, Xia, Cai, Wang, Kim, and
  Shibasaki]{Jiang19}
R.~Jiang, X.~Song, D.~Huang, X.~Song, T.~Xia, Z.~Cai, Z.~Wang, K.~Kim, and
  R.~Shibasaki.
\newblock Deep{U}rban{E}vent: {A} {S}ystem for {P}redicting {C}itywide {C}rowd
  {D}ynamics at {B}ig {E}vents.
\newblock In \emph{Proceedings of the 25th ACM SIGKDD International Conference
  on Knowledge Discovery and Data Mining}, 2019.

\bibitem[Karim et~al.(2019)Karim, Majumdar, Darabi, and Harford]{Karim19}
F.~Karim, S.~Majumdar, H.~Darabi, and S.~Harford.
\newblock Multivariate {LSTM-FCN}s for {T}ime {S}eries {C}lassification.
\newblock \emph{Neural Networks}, 116:\penalty0 237--245, 2019.

\bibitem[Karlsson et~al.(2016)Karlsson, Papapetrou, and
  Bostr{\"o}m]{Karlsson16}
I.~Karlsson, P.~Papapetrou, and H.~Bostr{\"o}m.
\newblock Generalized {R}andom {S}hapelet {F}orests.
\newblock \emph{Data Mining and Knowledge Discovery}, 30\penalty0 (5):\penalty0
  1053--1085, 2016.

\bibitem[Karlsson et~al.(2020)Karlsson, Rebane, and Gionis]{Karlsson20}
I.~Karlsson, J.~Rebane, and P.~Papapetrou~A. Gionis.
\newblock Locally and {G}lobally {E}xplainable {T}ime {S}eries {T}weaking.
\newblock \emph{Knowledge and Information Systems}, 62:\penalty0 1671–1700,
  2020.

\bibitem[Kotsiantis and Pintelas(2005)]{Kotsiantis05}
S.~Kotsiantis and P.~Pintelas.
\newblock Combining {B}agging and {B}oosting.
\newblock \emph{International Journal of Computational Intelligence},
  1\penalty0 (8):\penalty0 372--381, 2005.

\bibitem[Li et~al.(2018)Li, Rong, Meng, Lu, Kwok, and Cheng]{Li18}
J.~Li, Y.~Rong, H.~Meng, Z.~Lu, T.~Kwok, and H.~Cheng.
\newblock T{ATC}: {P}redicting {A}lzheimer’s {D}isease with {A}ctigraphy
  {D}ata.
\newblock In \emph{Proceedings of the 24th ACM SIGKDD International Conference
  on Knowledge Discovery and Data Mining}, 2018.

\bibitem[Lipton(2016)]{Lipton16}
Z.~Lipton.
\newblock The {M}ythos of {M}odel {I}nterpretability.
\newblock In \emph{Proceedings of the ICML Workshop on Human Interpretability
  in Machine Learning}, 2016.

\bibitem[Liu and Yao(1999)]{Liu99}
Y.~Liu and X.~Yao.
\newblock Ensemble {L}earning via {N}egative {C}orrelation.
\newblock \emph{Neural Networks}, 12\penalty0 (10):\penalty0 1399--1404, 1999.

\bibitem[Lundberg and Lee(2017)]{Lundberg17}
S.~Lundberg and S.~Lee.
\newblock A {U}nified {A}pproach to {I}nterpreting {M}odel {P}redictions.
\newblock In \emph{Proceedings of the 31st International Conference on Neural
  Information Processing Systems}, 2017.

\bibitem[Masoudnia and Ebrahimpour(2014)]{Masoudnia14}
S.~Masoudnia and R.~Ebrahimpour.
\newblock Mixture of {E}xperts: a {L}iterature {S}urvey.
\newblock \emph{Artificial Intelligence Review}, 42\penalty0 (2):\penalty0
  275--293, 2014.

\bibitem[Miller(2019)]{Miller19}
T.~Miller.
\newblock Explanation in {A}rtificial {I}ntelligence: {I}nsights from the
  {S}ocial {S}ciences.
\newblock \emph{Artificial Intelligence}, 267:\penalty0 1–38, 2019.

\bibitem[Pedregosa et~al.(2011)Pedregosa, Varoquaux, Gramfort, Michel, Thirion,
  Grisel, Blondel, Prettenhofer, Weiss, Dubourg, Vanderplas, Passos,
  Cournapeau, Brucher, Perrot, and Duchesnay]{scikit-learn}
F.~Pedregosa, G.~Varoquaux, A.~Gramfort, V.~Michel, B.~Thirion, O.~Grisel,
  M.~Blondel, P.~Prettenhofer, R.~Weiss, V.~Dubourg, J.~Vanderplas, A.~Passos,
  D.~Cournapeau, M.~Brucher, M.~Perrot, and E.~Duchesnay.
\newblock Scikit-{L}earn: {M}achine {L}earning in {P}ython.
\newblock \emph{Journal of Machine Learning Research}, 2011.

\bibitem[Ransbotham et~al.(2019)Ransbotham, Khodabandeh, Fehling, LaFountain, ,
  and Kiron]{Ransbotham19}
S.~Ransbotham, S.~Khodabandeh, R.~Fehling, B.~LaFountain, , and D.~Kiron.
\newblock Winning {W}ith {AI}.
\newblock In \emph{MIT Sloan Management Review and Boston Consulting Group},
  2019.

\bibitem[Ribeiro et~al.(2016)Ribeiro, Singh, and Guestrin]{Ribeiro16}
M.~Ribeiro, S.~Singh, and C.~Guestrin.
\newblock “{W}hy {S}hould {I} {T}rust {Y}ou?”: {E}xplaining the
  {P}redictions of {A}ny {C}lassifier.
\newblock In \emph{Proceedings of the 22nd ACM SIGKDD International Conference
  on Knowledge Discovery and Data Mining}, 2016.

\bibitem[Ribeiro et~al.(2018)Ribeiro, Singh, and Guestrin]{Ribeiro18}
M.~Ribeiro, S.~Singh, and C.~Guestrin.
\newblock Anchors: {H}igh-{P}recision {M}odel-{A}gnostic {E}xplanations.
\newblock In \emph{Proceedings of the 32nd AAAI Conference on Artificial
  Intelligence}, 2018.

\bibitem[Rudin(2019)]{Rudin19}
C.~Rudin.
\newblock Stop {E}xplaining {B}lack {B}ox {M}achine {L}earning {M}odels for
  {H}igh {S}takes {D}ecisions and {U}se {I}nterpretable {M}odels {I}nstead.
\newblock \emph{Nature Machine Intelligence}, 1:\penalty0 206--215, 2019.

\bibitem[Sch\"{a}fer and H\"{o}gqvist(2012)]{Schafer12}
P.~Sch\"{a}fer and M.~H\"{o}gqvist.
\newblock {SFA}: {A} {S}ymbolic {F}ourier {A}pproximation and {I}ndex for
  {S}imilarity {S}earch in {H}igh {D}imensional {D}atasets.
\newblock In \emph{Proceedings of the 15th International Conference on
  Extending Database Technology}, pages 516--527, 2012.

\bibitem[Sch{\"a}fer and Leser(2017)]{Schafer17}
P.~Sch{\"a}fer and U.~Leser.
\newblock Multivariate {T}ime {S}eries {C}lassification with {WEASEL+MUSE}.
\newblock \emph{arXiv}, 2017.

\bibitem[Schapire(1990)]{Schapire99}
R.~Schapire.
\newblock The {S}trength of {W}eak {L}earnability.
\newblock \emph{Machine Learning}, 5:\penalty0 197--227, 1990.

\bibitem[Selvaraju et~al.(2019)Selvaraju, Das, Vedantam, Cogswell, Parikh, and
  Batra]{Selvaraju19}
R.~Selvaraju, A.~Das, R.~Vedantam, M.~Cogswell, D.~Parikh, and D.~Batra.
\newblock Grad-{CAM}: {V}isual {E}xplanations from {D}eep {N}etworks via
  {G}radient-{B}ased {L}ocalization.
\newblock \emph{International Journal of Computer Vision}, 128:\penalty0
  336--359, 2019.

\bibitem[Sesmero et~al.(2015)Sesmero, Ledezma, and Sanchis]{Sesmero15}
M.~Sesmero, A.~Ledezma, and A.~Sanchis.
\newblock Generating {E}nsembles of {H}eterogeneous {C}lassifiers {U}sing
  {S}tacked {G}eneralization.
\newblock \emph{Wiley Interdisciplinary Reviews: Data Mining and Knowledge
  Discovery}, 5\penalty0 (1):\penalty0 21--34, 2015.

\bibitem[Seto et~al.(2015)Seto, Zhang, and Zhou]{Seto15}
S.~Seto, W.~Zhang, and Y.~Zhou.
\newblock Multivariate {T}ime {S}eries {C}lassification {U}sing {D}ynamic
  {T}ime {W}arping {T}emplate {S}election for {H}uman {A}ctivity {R}ecognition.
\newblock In \emph{Proceedings of the 2015 IEEE Symposium Series on
  Computational Intelligence}, 2015.

\bibitem[Sharkey and Sharkey(1997)]{Sharkey97}
A.~Sharkey and N.~Sharkey.
\newblock Combining {D}iverse {N}eural {N}ets.
\newblock \emph{The Knowledge Engineering Review}, 12\penalty0 (3):\penalty0
  231--247, 1997.

\bibitem[Shokoohi-Yekta et~al.(2017)Shokoohi-Yekta, Hu, Jin, Wang, and
  Keogh]{Shokoohi17}
M.~Shokoohi-Yekta, B.~Hu, H.~Jin, J.~Wang, and E.~Keogh.
\newblock Generalizing {DTW} to the {M}ulti-{D}imensional {C}ase {R}equires an
  {A}daptive {A}pproach.
\newblock \emph{Data Mining and Knowledge Discovery}, 31:\penalty0 1--31, 2017.

\bibitem[Tuncel and Baydogan(2018)]{Tuncel18}
K.~Tuncel and M.~Baydogan.
\newblock Autoregressive {F}orests for {M}ultivariate {T}ime {S}eries
  {M}odeling.
\newblock \emph{Pattern Recognition}, 73:\penalty0 202--215, 2018.

\bibitem[Wang et~al.(2017)Wang, Yan, and Oates]{Wang17}
Z.~Wang, W.~Yan, and T.~Oates.
\newblock Time {S}eries {C}lassification from {S}cratch with {D}eep {N}eural
  {N}etworks: {A} {S}trong {B}aseline.
\newblock In \emph{Proceedings of the 2017 International Joint Conference on
  Neural Networks}, 2017.

\bibitem[Wistuba et~al.(2015)Wistuba, Grabocka, and
  Schmidt{-}Thieme]{Wistuba15}
M.~Wistuba, J.~Grabocka, and L.~Schmidt{-}Thieme.
\newblock Ultra-{F}ast {S}hapelets for {T}ime {S}eries {C}lassification.
\newblock \emph{arXiv}, 2015.

\bibitem[Wolpert(1996)]{Wolpert96}
D.~Wolpert.
\newblock The {L}ack of {A} {P}riori {D}istinctions {B}etween {L}earning
  {A}lgorithms.
\newblock \emph{Neural Computation}, 8\penalty0 (7):\penalty0 1341–1390,
  1996.

\bibitem[Zerveas et~al.(2021)Zerveas, Jayaraman, Patel, Bhamidipaty, and
  Eickhoff]{Zerveas21}
G.~Zerveas, S.~Jayaraman, D.~Patel, A.~Bhamidipaty, and C.~Eickhoff.
\newblock A {T}ransformer-{B}ased {F}ramework for {M}ultivariate {T}ime
  {S}eries {R}epresentation {L}earning.
\newblock In \emph{Proceedings of the 27th ACM SIGKDD Conference on Knowledge
  Discovery and Data Mining}, 2021.

\bibitem[Zhang(2004)]{Zhang04}
H.~Zhang.
\newblock The {O}ptimality of {N}a\"ive {B}ayes.
\newblock In \emph{Proceedings of the 17th Florida Artificial Intelligence
  Research Society Conference}, 2004.

\bibitem[Zhang et~al.(2020)Zhang, Gao, Lin, and Lu]{Zhang20}
X.~Zhang, Y.~Gao, J.~Lin, and C.~Lu.
\newblock Tap{N}et: {M}ultivariate {T}ime {S}eries {C}lassification with
  {A}ttentional {P}rototypical {N}etwork.
\newblock In \emph{Proceedings of the 34th {AAAI} Conference on Artificial
  Intelligence}, 2020.

\bibitem[Zou and Hastie(2005)]{Zou05}
H.~Zou and T.~Hastie.
\newblock Regularization and {V}ariable {S}election via the {E}lastic {N}et.
\newblock \emph{Journal of the Royal Statistical Society. Series B (Statistical
  Methodology)}, 67\penalty0 (2):\penalty0 301--320, 2005.

\end{thebibliography}

\end{document}